\documentclass{article} 
\usepackage[preprint,nonatbib]{neurips_2022}

\usepackage{amsthm}
\theoremstyle{plain}
\theoremstyle{definition}

\theoremstyle{remark}

\usepackage{wrapfig,lipsum,booktabs}

\usepackage{xcolor}

\usepackage{subcaption}
\usepackage{graphicx,amssymb,amsmath,mathtools,amsthm,mathrsfs}

\usepackage[hidelinks]{hyperref}
\usepackage{url}
\usepackage{cancel}
\definecolor{cite_color}{HTML}{114083}
\definecolor{link_color}{RGB}{0,102,102}
\definecolor{link_color}{RGB}{153, 0,0}  
\definecolor{url_color}{RGB}{153, 102,  0}
\definecolor{emp_color}{RGB}{0,0,255}
\hypersetup{
 colorlinks,
 citecolor=cite_color,
 linkcolor=link_color,
 urlcolor=url_color}

 \usepackage{hyperref}

\usepackage[capitalize,noabbrev]{cleveref}

\usepackage{answers}
\usepackage{setspace}
\usepackage{graphicx}
\usepackage{graphics}
\usepackage{subcaption}
\usepackage{enumitem}
\usepackage{multicol}
\usepackage{mathrsfs}
\usepackage{bbm}
\usepackage{amsmath,amsthm,amssymb}
\usepackage{rotating}
\usepackage{float}
\usepackage{url}
\usepackage{mathtools}
\usepackage{caption}
\usepackage{algorithm}
\usepackage{algorithmic}
\usepackage[toc,page]{appendix}
\usepackage{color}
\usepackage{tikz}
\usetikzlibrary{bayesnet}
\usepackage{tabularx}
\usepackage{array}
\usepackage{multirow}
\usepackage{hyperref}

\usepackage{xcolor}

\DeclarePairedDelimiterX{\infdivx}[2]{\Big[}{\Big]}{%
  #1\;\delimsize\|\;#2%
}

\newcommand{\vecd}{\mathbf{d}}

\newcommand{\vectheta}{\pmb{\theta}}

\newcommand{\R}{\mathbbm{R}}

\newtheorem{Proposition}{Proposition}

\DeclareMathOperator*{\argmax}{argmax}

\title{Mutual Information Multinomial Estimation}

\author{
    Yanzhi Chen$^{1}$\thanks{Equal contribution. \enskip Code will be made available at github: \href{https://github.com/cyz-ai/neural-MI-estimate}{https://github.com/cyz-ai/neural-MI-estimate.}}, \enskip Zijing Ou$^{2*}$, Adrian Weller$^{1,3}$, Yingzhen Li$^{2}$\\
    \hspace{-0.25cm} $^1$University of Cambridge, $^2$Imperial College London, $^3$Alan Turing Institute\\
}

\begin{document}

\maketitle

\begin{abstract}
Estimating mutual information (MI) from data is a fundamental task in machine learning and data science, yet it remains highly challenging even with state-of-the-art estimators. This work proposes a new distribution-free MI estimator based on multinomial classification. Unlike existing works that only discern between the joint distribution and the marginal distribution, which can easily overfit in high-MI settings, our method compares them with extra reference distributions. These reference distributions share the same marginals as the original distributions but 
have known dependence structures, thereby providing additional signals for accurate dependency modelling. Experiments on synthetic tasks with non-Gaussian, high-dimensional data and real-world applications including Bayesian experimental design and self-supervised learning demonstrate the potential of our approach.
\end{abstract}

\section{Introduction}

Mutual information (MI), defined as the Kullback-Leibler divergence between the joint probability and the product of marginals, is a fundamental measure of the dependence between random variables. Unlike correlation, MI captures non-linear statistical dependencies between high-dimensional variables, making it a robust measure of  dependence in diverse cases \cite{kinney2014equitability}. Consequently, MI has been widely applied across diverse fields, such as representation learning \cite{van2018representation,hjelm2018learning,chen2020simple}, Bayesian experimental design \cite{kleinegesse2020bayesian,foster2021deep}, information bottleneck \cite{tishby2015deep,alemi2016deep}, domain generalization \cite{li2022invariant, gholami2020unsupervised}, and causality \cite{solo2008causality,kurutach2018learning}.

Despite its widespread applications, estimating MI in high-dimensional spaces from raw data is highly challenging.
Early methods work in a non-parametric fashion, utilizing techniques like binning or kernel density estimation \cite{fraser1986independent, darbellay1999estimation, kraskov2004estimating, moon1995estimation, kwak2002input}. As non-parameteric methods typically do not scale well with sample size or data dimensionality \cite{gao2015efficient}, recent studies have focused on utilizing powerful parametric models such as deep neural networks in MI estimation \cite{belghazi2018mutual, van2018representation, franzeseminde, huinfonet}. In these estimators, a neural network (often known as the critic) is trained to discern samples from the joint distribution and the marginal distribution. Upon convergence, it can be shown that the loss function of these estimators will optimize a variational lower bound of MI, thereby offering an estimate to MI. Compared to non-parametric methods, estimating MI through neural network training not only handles large sample sizes more effectively but also scales better with high-dimensional data. Thereby, it is increasingly gaining popularity in both machine learning and statistics communities.

Although neural network-based MI estimators have achieved success in numerous tasks, they are recently found to suffer from the so-called high-discrepancy issue, particularly when the underlying mutual information is high or the data dimensionality is high. In such scenarios, distinguishing samples from the joint and the marginal distributions becomes considerably easier, leading the network to easily overfit  with limited data. This curse of high-discrepancy has been empirically observed in multiple applications \cite{ozair2019wasserstein, srivastava2023estimating, chen2023learning} and has also been formally analyzed in theoretical studies \cite{mcallester2020formal, song2019understanding}.

In this work, we propose Mutual Information Multinomial Estimation (MIME), a novel approach tailored for estimating mutual information in high-discrepancy settings. Unlike many existing methods that solely focus on discerning samples between the joint  and the marginal distributions, which can easily overfit in high-discrepancy scenarios, MIME estimates MI by learning to compare a set of reference distributions that are constructed by generative modelling. 
These distributions maintain the same marginal distributions but exhibit various joint dependency structures, thereby providing rich training signals that encourages the network to learn a useful representation for accurate dependence modelling. To summarize, our main contributions are:
\begin{itemize}[leftmargin=*]
    \item We propose a new mutual information estimator tailored for high-MI cases, combining the strengths of existing generative and distribution-free methods while avoiding their weaknesses;
    \item We systematically evaluate our estimator on high-dimensional, non-Gaussian synthetic dataset with known MI as well as real-world applications such as Bayesian experimental design and SSL.
\end{itemize}

\section{Background}
\label{sec:bg}

The mutual information (MI) between variables $X$ and $Y$ is defined as the Kullback-Leibler (KL) divergence between the joint distribution $p(x, y)$ and the product of the marginal  $p(x)p(y)$
\begin{equation}
      I(X; Y) = KL[p(x, y)\|p(x)p(y)] = \mathbb{E}\left[ \log \frac{p(x,y)}{p(x)p(y)} \right].
      \label{eq:mine}
\end{equation}
It is obvious that stronger dependence between $X$ and $Y$ leads to larger divergence between $p(x, y)$ and $p(x)p(y)$. 
Thereby, MI can measure the dependence between two random variables.
Given a dataset $\mathcal{D} = \{x^{(i)}, y^{(i)} \}$ drawn from the joint distribution $p(x,y)$, our goal in this work is to estimate $I(X; Y)$ from the empirical samples $\mathcal{D}$.
Before delving into our approach, we first revisit some existing MI estimators and their limitations.

\textbf{Distribution-free MI estimator}. Many methods have been developed to estimate \eqref{eq:mine} from $\mathcal{D}$ without making any assumptions about the underlying probability distributions. One popular method is to utilize the Donsker-Varadhan (DV) representation of KL divergence \cite{donsker1983asymptotic}:
\begin{equation}
     I(X;Y) \geq \sup_{f \in \mathcal{F}} \mathbb{E}_{ p(x, y)}[f(x, y)] - \log \mathbb{E}_{p(x)p(y)}[e^{f(x, y)}] =: \hat{I}_{\text{DV}}(X; Y),
    \label{eq:dv}
\end{equation}
where $\mathcal{F}$ can be any class of function $\mathcal{F}: \mathcal{X} \rightarrow \mathbb{R}$ satisfying the integrability constraints of the theorem. A typical choice of $\mathcal{F}$ is the class of deep neural network. In such case, MI estimate reduces to training a neural network $f$ to maximize the objective \eqref{eq:dv}. Similarly, the NWJ estimator \cite{nguyen2010estimating} makes use of the $f$-divergence lower bound to estimate MI:
$ I(X;Y) \geq \sup_{f \in \mathcal{F}} \mathbb{E}_{ p(x, y)}[f(x, y)] - \mathbb{E}_{p(x)p(y)}[e^{f(x, y)-1}] =: \hat{I}_{\text{NWJ}}(X; Y)$, which has a smaller variance than the DV estimator at the cost of being biased. Both DV and NWJ estimators leads to a lower bound of MI.

Alternatively, one can estimate MI by employing techniques from density ratio estimate~\cite{rhodes2020telescoping, mukherjee2020ccmi, srivastava2023estimating}:
\begin{equation}
     I(X;Y) \approx \mathbb{E}_{p(x, y)}[\log \hat{r}(x, y)] =: \hat{I}_{\text{ratio}}(X; Y)
    \label{eq:ratio}
\end{equation}
where $\hat{r}(x, y) \approx p(x, y)/p(x)p(y)$ is the estimated density ratio. To learn $r$, one can adopt a process similar to GAN training~
\cite{goodfellow2014generative, nowozin2016f}, where a classifier is trained to distinguish samples from the joint distribution $p(x, y)$ and samples from the marginal distribution $p(x)p(y)$. It can be shown that upon convergence, the output of the classifier estimates the log density ratio. This type of estimator is found more robust than the above DV-like estimators~\cite{hjelm2018learning} and is widely used in information-theoretic representation learning~\cite{hjelm2018learning, chen2020neural}. However, the MI estimated in this way is no more a lower bound.

\textbf{Curse of high discrepancy}. Despite their conceptual advantages compared to non-parametric methods, distribution-free estimators often suffer from the so-called high-discrepancy issue in practical use. High-discrepancy issue typically happens when the dimensionality of data is high and/or the underlying MI is high, which results in a large KL between $p(x, y)$ and $p(x)p(y)$. However, it is shown that asymptotically, any distribution-free estimator of the KL divergence between two distributions $p$ and $q$ will inevitably suffer from an estimation variance that is exponential to the ground truth value~\cite{song2019understanding, mcallester2020formal}:
\begin{equation}
    \mathbb{V}[\hat{KL}[p\|q]] \approx O(e^{KL[p\|q]})
\end{equation}
which implies that methods based on KL divergence estimation (e.g. DV, NWJ) will be impractical if the underlying MI is large or the intrinsic dimensionality of the data is high, both of which can lead to a large KL. The situation becomes even worse when we learn estimators with mini-batches. This issue raises concerns about the reliability of DV- and NWJ-like estimators in high-MI settings.

We highlight that the impact of the high discrepancy issue is well beyond KL divergence-based estimators. For example,  the MI estimated by the well-known InfoNCE estimator~\cite{chen2020simple, oord2018representation} is shown to be bounded by the sample size $S$~\cite{van2018representation}:
\begin{equation}
    \hat{I}_{\text{InfoNCE}}(X; Y) \leq \log S
\end{equation}
which implies InfoNCE will largely underestimate MI under small batch size if the true MI is high. Furthermore, recent works have confirmed that high-discrepancy issue also exists in ratio-based estimator \eqref{eq:ratio}, where the classifier tends to overfit when samples $x, y \sim p(x, y)$ and samples $x, y \sim p(x)p(y)$ are overly easy to discern~\cite{rhodes2020telescoping, choi2021featurized, chen2023learning, srivastava2023estimating, gruber2024overcoming}. One intuitive explanation is that when the gap between $p(x, y)$ and $p(x)p(y)$ is large, there exist many classifiers that can achieve almost perfect  accuracy but have differently estimated density ratios, leading to inaccurate estimate of MI.

\section{Method}

In this work, we propose a new MI estimator aiming to address the aforementioned high-discrepancy issue in distribution-free MI estimation. The key idea is, unlike existing works that solely compare the joint distribution $p(x, y)$ with the product of the marginal distributions $p(x)p(y)$, which easily causes overfitting, we also compare them with two reference distributions whose marginals are (almost) the same as $p(x, y)$ and $p(x)p(y)$ but have Gaussian dependence structures. These additional comparisons provide more fine-grained  signals for  accurately modelling the dependence structures of $p(x, y)$ and $p(x)p(y)$, thereby allowing us to better model the dependence between $X$ and $Y$.

\textbf{Reference-based MI estimation}. The core of our method is to formulate MI estimation as a multinomial classification problem, where we not only distinguish samples from $p(x, y)$, $p(x)p(y)$ but also samples from reference distributions. Specifically, consider classifying samples $x, y$ from four distributions $p_1(x, y), p_2(x, y), p_3(x, y), p_4(x, y)$ with a classifier $h$. This can be done by training $h$ with the following objective function:
\begin{equation}
\begin{aligned}
    \max_h \enskip \mathcal{L}(h) = \mathbb{E}_{p(c)p_c(x, y)}&\Big[\log \frac{e^{h_c(x, y)}}{\sum^K_k e^{h_k(x, y)}}\Big], \\
    p_1(x, y) = p(x, y), \quad\enskip p_2(x, y) = q(x, y), \quad\enskip  p_3&(x, y) = q(x)q(y), \quad\enskip p_4(x, y) = p(x)p(y).
\end{aligned}
\label{eq:obj}
\end{equation}
where $h_c$ denotes the logit (i.e. the unnormalized class probability) computed for class $c$ and $q(x, y)$ is a reference distribution. A crucial property of $q(x, y)$ is that it is \emph{marginal preserving}, that its marginal distributions $q(x)$ and $q(y)$ are almost the same as that of $p(x, y)$ but have a known dependence structure (e.g. Gaussian). The marginal-preserving property of $q$ is motivated by the inductive bias that MI is irrelevant to marginals, so a good reference distribution should only differ from $p(x, y)$ in their dependence structures. The details of how to construct $q(x, y)$ will be clear soon. 

Upon convergence, it can be shown that when the prior $p(c)$ is uniform, the optimal $h$ satisfies~\cite{srivastava2023estimating}:
\begin{equation}
    \log \frac{p_i(x, y)}{p_j(x, y)} = h_i(x, y) - h_j(x, y),
    \label{eq:ratio_mime}
\end{equation}
thus allowing the mutual information to be estimated as:
\begin{equation}
    I(X;Y) \approx \hat{I}(X;Y | h) := \mathbb{E}_{p(x, y)}\Big[h_1 (x, y) - h_4 (x, y)\Big],
    \label{eq:mi-est-ours}
\end{equation}
where the expectation on the RHS can be readily approximated by Monte Carlo integration.

We provide an intuitive explanation of why the above method for estimating $I(X; Y)$ is more advantageous than previous methods. In previous methods, samples from $p(x, y)$ are only compared with those from $p(x)p(y)$. Such a task is too simple and can easily overfit if $p(x, y)$ and $p(x)p(y)$ are vastly different from each other; see Section \ref{sec:bg}. Now, by additionally comparing them with two reference distributions whose marginals are the same as $p(x, y)$ and $p(x)p(y)$ and the dependence structures approximate that of $p(x, y)$ and $p(x)p(y)$, the network has to learn a useful representation to distinguish the small difference in dependence structures, thereby effectively avoiding overfitting.

\textbf{Constructing reference distributions}. We next discuss how to construct the reference distribution $q(x, y)$ whose marginals are close to $p(x, y)$ but with a known dependence structure. In this work, we design $q$ to be a \emph{vector Gaussian copula}. A vector Gaussian copula  can either be seen as a special case of the recent vector copula model~\cite{fan2023vector} or a generalization of the classic Gaussian copula model.

Specifically, data $x, y \sim q(x, y)$ from a vector Gaussian copula can be seen as generating from the following data generation process:
\begin{equation}
\begin{aligned}
    x = f(\epsilon_{\leq d}), & \quad y = g(\epsilon_{>d}) \\
     \epsilon \sim \mathcal{N}(&\epsilon; 0, \Sigma)
\end{aligned}
\label{eq:GGC}
\end{equation}
where $f: \R^d \to \R^d$ and $g: \R^{d'} \to \R^{d'}$ are two bijective functions and $\Sigma \in \R^{(d+d') \times (d+d')}$ is a p.s.d matrix where $\Sigma_{ii} = 1, \forall i$. Here $\epsilon_{\leq d}$ and $\epsilon_{>d}$ are the first $d$ and the last $d'$ dimensions of $\epsilon$. Note that this model degenerates to a Gaussian copula when $f$ and $g$ are element-wise  functions\footnote{We adopt this element-wise implementation when applying our method to representation learning tasks. }. 

We implement the two bijective functions $f$ and $g$ by flow-based models e.g.~\cite{papamakarios2017masked, lipman2022flow}. Under this setup, we learn the parameters $\{f, g, \Sigma\}$ by first learning $f$ and $g$ with data $x \sim p(x)$ and $y \sim p(y)$ from the two marginals, then learning $\Sigma$ as $\Sigma = \mathbb{E}[\epsilon\epsilon^{\top}]$ where $\epsilon_{\leq d} = f^{-1}(x)$ and  $\epsilon_{>d} = g^{-1}(y)$.  

The above vector Gaussian copula modelling has several advantages. (a) It satisfies the 
\emph{marginal preserving} property well: compared to modelling $q$ as a single flow-based model, the marginals in vector Gaussian copula can often be learned more accurately due to the reduced dimensionality in density estimate and the separate training of the two marginals. This guarantees that the marginal distributions of $q(x, y)$ would be very close to that of $p(x, y)$; (b) It is \emph{moderately close to} $p(x, y)$, that it is much closer to $p(x, y)$ than $p(x)p(y)$ but is not identical to $p(x, y)$, making it a useful reference for providing effective signals for understanding the true dependence structure of $p(x, y)$.

The whole process of our MI estimator can be seen as that we compare the dependence structures of $p(x, y)$ and $p(x)p(y)$ with Gaussian dependence structures, which provides additional information.

\section{Theoretic analysis}
\label{sec:theory}

In practice, we replace the expectation in \eqref{eq:obj} with $N$ samples to learn our MIME estimator via: $\hat{h} = \arg\max_h \mathcal{L}_N (h)$. After training, one can estimate the mutual information via Monte Carlo estimation: $\hat{I}_N (X;Y|\hat{h}) := \frac{1}{N} \sum_{i=1}^N \hat{h}_1 (x_i, y_i) - \hat{h}_4 (x_i, y_i)$.
In \cref{consistent-of-mime}, we prove that our MIME estimator is consistent. \\

\begin{Proposition} \label{consistent-of-mime}
(\emph{Consistency of multinomial MI estimate}). Assuming that the multinomial classifier $h_c: X \times Y \rightarrow \mathbb{R}$ is uniformly bounded. Then, for every  $\varepsilon > 0$ there exists $N(\varepsilon) \in \mathbb N$, such that 
\[
    |\hat{I}_{N}(X; Y|\hat{h}) - I(X; Y)| < \varepsilon, \forall N \geq N(\varepsilon), a.s..
\]
\end{Proposition}
\emph{Proof}. See Appendix A.  \qed \\

This provides a desirable justification for our approach, confirming that the probability that the estimated MI of MIME becomes arbitrarily close to the true MI converges to one, as the amount of data used in the estimation approaches infinity. 

Furthermore, we have the following formal result about the advantages of our proposed estimator. \\

\begin{Proposition}
(\emph{Controlled error in multinomial MI estimate}). Define $\log \hat{r}_{i, j}(x, y)  \coloneqq h_i(x, y) - h_j(x, y)$ with $h$ being the classifier defined as above. Upon convergence, we have that:
\[
    \Big |\log \hat r_{1, 4}(x, y) - \log \frac{p(x, y)}{p(x)p(y)} \Big | \leq \lambda { \sup_i} \Big |\log \hat r_{i, i+1}(x, y) - \log \frac{p_{i}(x, y)}{p_{{i}+1}(x, y)} \Big| ,
\]
where $\lambda \leq 3$ is some real-valued positive constant. 
\end{Proposition}
\emph{Proof}. See Appendix A.  \qed \\

In other words, if the ratios $p(x, y)/q(x, y)$, $p(x)p(y)/q(x)q(y)$ and $q(x, y)/q(x)q(y)$ all estimated accurately in $h$, then the ratio $p(x,y)/p(x)p(y)$ estimated by the same network will also be accurate. This is exactly the case in our method: the first two ratios are by construction easy to estimate due to the reduced distributional discrepancy. For the ratio $q(x, y)/q(x)q(y)$, the discrepancy between the two distributions may be high, but we can sample infinitely many samples from these two distributions during mini-batch learning, so the ratio between them can also be learned accurately.

\section{Related works}

\textbf{Alleviating high-discrepancy issue in  statistical divergence estimation}. Several methods have been developed to address the high discrepancy issues in the estimation of statistical divergence the estimation of density ratio. One method is to reduce the estimation variance by clipping the output of the estimator~\cite{song2019understanding}, which can be seen as trading low variance with additional bias. Another line of methods solves the problem by the divide-and-conquer principle~\cite{rhodes2020telescoping, choi2022density}, where the statistical divergence/density ratio between some intermediate distributions are first estimated, followed by an aggregation step to recover the original statistical divergence/density ratio from these intermediate estimations. Due to the reduced distributional discrepancy, each of these intermediate estimates can be expected to be accurate. However, divide-and-conquer methods are found to suffer from the \emph{distribution-shift} issue during the aggregation step, which can lead to  estimation inaccuracy. Unlike these methods, our method does not incur extra estimation bias and is free from  distribution shift.

\textbf{Multi-classes ratio estimate}. Similar to our work, some recent works also utilize a set of reference distributions $q$ to improve density ratio estimate \cite{yu2021unified,srivastava2023estimating}. A crucial difference between our method and these approaches is that the reference distributions in our method are \emph{marginal-preserving} i.e. their marginal distributions $q(x)$ and $q(y)$ are close to the original marginal distributions. As aforementioned, such marginal-preserving property is important as it enforces the network to focus on modelling the changes in dependence structures rather than modelling the marginal distributions, which are irrelevant for MI estimation. This importance will be verified in the experiment section.

\textbf{Generative methods for mutual information estimation}.  In addition to the above distribution-free methods for MI estimate, 
there also exist a set of generative MI estimators~\cite{mcallester2020formal,cheng2020club,butakov2024mutual} based on density estimation. These generative methods work by directly approximating  densities by some generative models e.g. a flow-based model~\cite{papamakarios2017masked,lipman2022flow} rather than comparing between densities, thereby are free from the high-discrepancy issue. However, the performance of such estimators is dominated by the quality of the estimated densities, and density estimate as done by flow-based model is well-known difficult in high-dimensional cases~\cite{rippel2013highdimensional,Oord2016Pixel,chen2020neural,lipman2022flow} due to the complex dependence among high-dimensional data and potential model mis-specification. Our method also makes use of flow-based models, but we only use them as high-quality reference distributions, which is a safer strategy.

\section{Experiment}
\textbf{Baselines}. We compare our new multinomial MI estimator with four representative baselines in the field: MINE \cite{belghazi2018mutual}, InfoNCE \cite{van2018representation}, MRE \cite{srivastava2023estimating} and DoE \cite{mcallester2020formal}. Specifically:

\begin{itemize}[leftmargin=*]
    \item \emph{Mutual Information Neural Estimate (MINE)}. This  method estimates MI by the Donsker-Varadhan representation and a neural network $f$: $I(X;Y) = \sup_{f} \mathbb{E}_{p(x, y)}[f(x, y)] - \log \mathbb{E}_{p(x)p(y)}[e^{f(x, y)}]$.
    \item \emph{InfoNCE}. Let $Y = \{y_1, \dots, y_{N-1}\}$ be $N-1$ negative samples from a proposal distribution, \emph{InfoNCE} estimates MI by learning a neural network $f$ to minimize the loss function $\mathcal{L} = -E_{p(x,y)} \left[ \log \frac{f(x,y)}{\sum_{y_i \in Y \cup \{y\}} f(x, y_i)} \right]$, which is upper bounded by MI: $I(X,Y) \geq \log(N) - \mathcal{L}$.
    \item \emph{Multinomial Ratio Estimate (MRE)}. A state-of-the-art method for density ratio estimation which also works by classifying samples from multiple (reference) distributions. The key difference between MRE and our method is that the  reference distributions in MRE are not marginal-preserving.
    \item \emph{Difference of Entropy (DoE)}. A generative method which works by first estimating the two entropies $H[Y]$ and $H[X|Y]$ with density estimators $q$, then calculate MI as $I(X; Y) = H[Y] - H[X|Y]$. Here, we study an improved variant of DoE where $q$ is realized as a flow model~\cite{papamakarios2017masked}. 
\end{itemize}

\textbf{Neural network architecture and optimizer}. For fair comparison, we use the same network architecture for the critic $f(x, y)$ in all the MI estimators considered. This network takes the form of a MLP with 3 hidden layers,  each of which has 500 neurons. A densenet architecture \cite{huang2017densely} is used for the network, where we concatenate the input of the first layer (i.e., $x$ and $y$) and the representation of the penultimate layer before feeding them to the last layer. Leaky ReLU \cite{maas2013rectifier} is used as the activation function for all hidden layers. We train all networks by Adam~\cite{kingma2014adam} with its default settings, where the learning rate is set to be $5\times10^{-4}$ and the batch size is set to be $512$. Early stopping and a slight weight decay ($1\times10^{-6}$) are applied to avoid overfitting. No dropout or batch normalization is used.

\textbf{Computation resource}. Each run of the experiments below is conducted with a NVIDIA A10 GPU.

\subsection{Synthetic data}
\label{section:synthetic}
We first investigate the performance of our method on tractable simulated cases, where the mutual information is either analytically known or can be computed numerically up to very high precision. 

\textbf{Setup}. We consider estimating MI with data simulated from the following two classes of models, both of which are more complex than the benchmarks considered in existing works~\cite{belghazi2018mutual, cheng2020club, huinfonet}:

\begin{itemize}[leftmargin=*]
    \item \emph{Non-linear transformation of multivariate Gaussian}. The first model we consider is a generalized, high-dimensional version of the cubic Gaussian tasks widely used in literature~\cite{belghazi2018mine, song2019understanding, cheng2020club}. Data $X \in \R^d, Y \in \R^d$ in this task is generated as follows:
\[
    x = f(\epsilon_{\leq d}), \qquad y = g(\epsilon_{>d})
 , \qquad \epsilon \sim \mathcal{N}(\epsilon; \mathbf{0}, \Sigma),
\]
where $\Sigma$ is a sparse convariance matrix satisfying $\Sigma_{i,i+d} = \rho, \Sigma_{i,i} = 1$ and $\Sigma_{i, j} = 0$ for any other $j \neq i+d, j\neq i$. $f: \R^{d} \to \R^{d}$ and $g: \R^{d} \to \R^{d}$ are some bijective functions (e.g. $f(x) = \mathbf{A}\tanh(x)$ with $\mathbf{A}$ being an invertible matrix).  Here $\rho$ controls the dependence between $X$ and $Y$. The ground truth MI in this model is given by $I(X; Y) = I(\epsilon_{\leq d}; \epsilon_{>d}) = -\frac{d}{2} \log (1-\rho^2)$. 

\item \emph{Mixture of multivariate Gaussians (MoG)}. The second case we consider is the mixture of $M$ high-dimensional Gaussian distributions whose likelihood function being is as follows:
\[
    p(x, y) = \frac{1}{M} \sum^M_k \mathcal{N}([x, y]; \mu_k, \Sigma_k)
\]
where $\Sigma_1...\Sigma_M$ is a set of sparse covariance matrices that are the same as in the non-linear Gaussian model. Note that the dependence factor $\rho_k$ can be different for each $\Sigma_k$. Unlike the non-linear Gaussian model, the MI of this model is not analytically known; however, we can approximate it with Monte Carlo integration due to the availability of both the joint distribution $p(x, y)$ and the marginal distribution $p(x)p(y)$. The variance of MC integration vanishes given sufficient data.
\end{itemize}

\begin{figure}[t!]
\centering
\hspace{-0.03\textwidth}
\begin{subfigure}{.255\textwidth}
\centering
\includegraphics[width=1.0\linewidth]{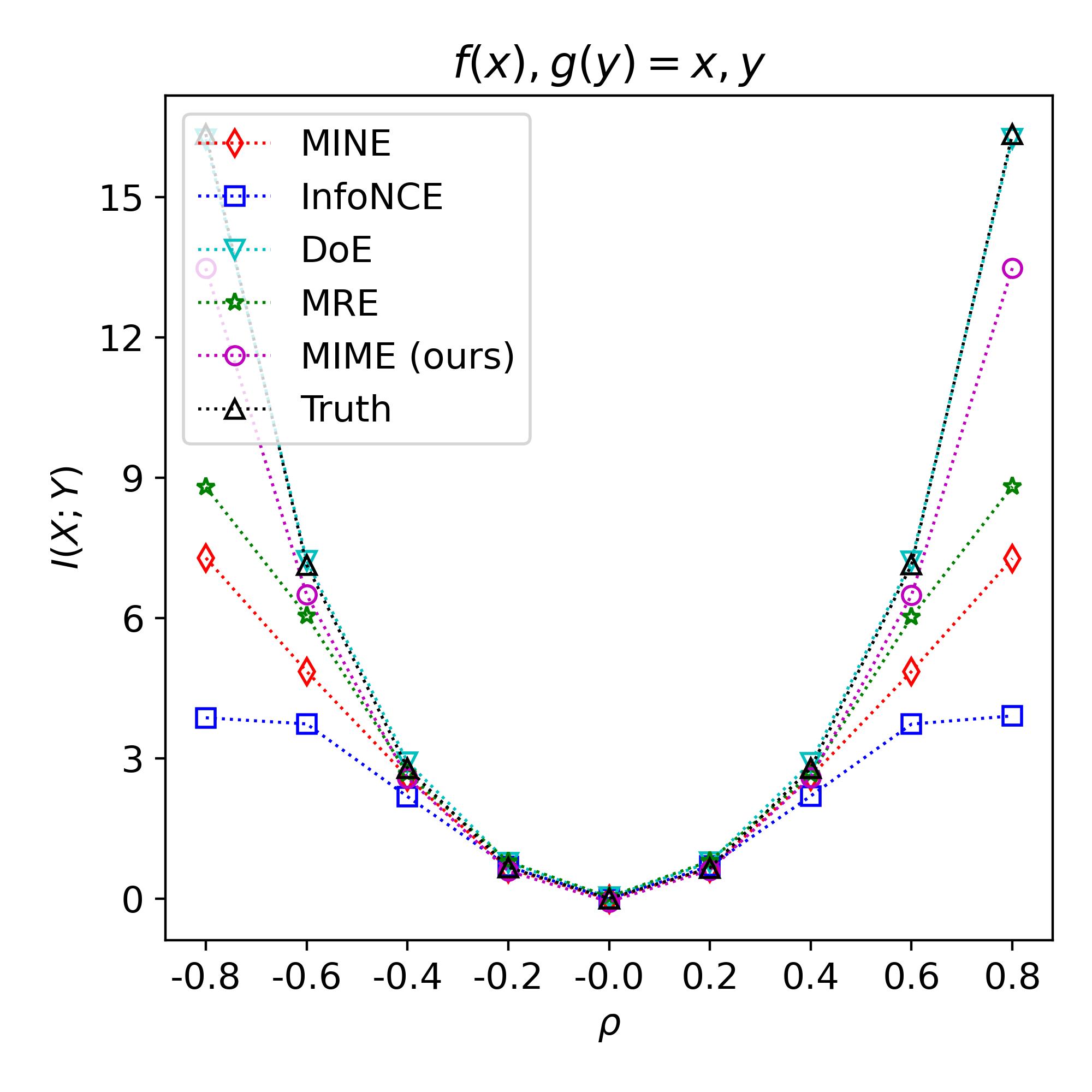}
\caption{\centering $x, y$}
\end{subfigure}
\hspace{-0.015\textwidth}
\begin{subfigure}{.255\textwidth}
\centering
\includegraphics[width=1.0\linewidth]{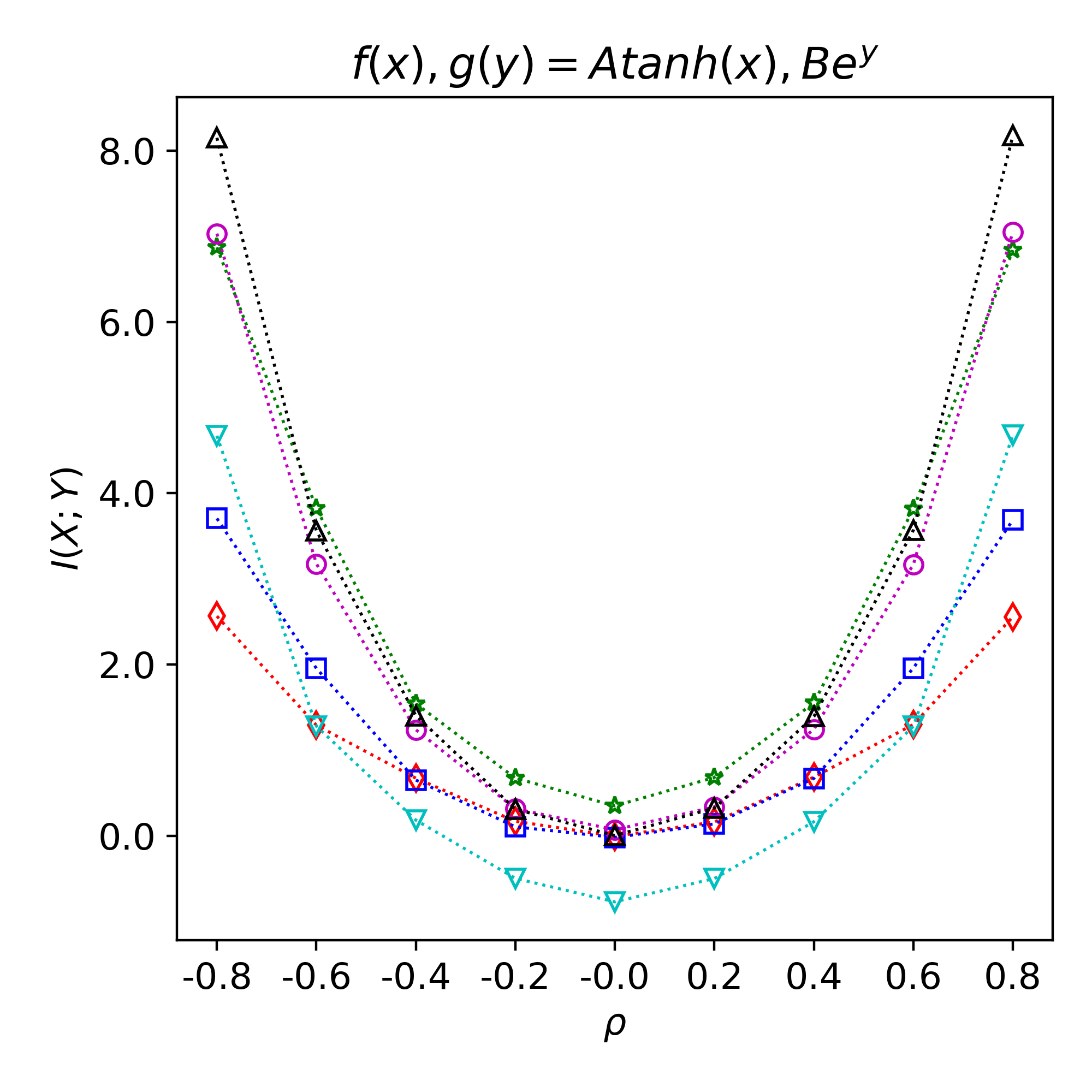}
\caption{\centering $\textbf{A}\text{tanh}(x), \textbf{B}e^y$}
\end{subfigure}
\hspace{-0.015\textwidth}
\begin{subfigure}{.255\textwidth}
\centering
\includegraphics[width=1.0\linewidth]{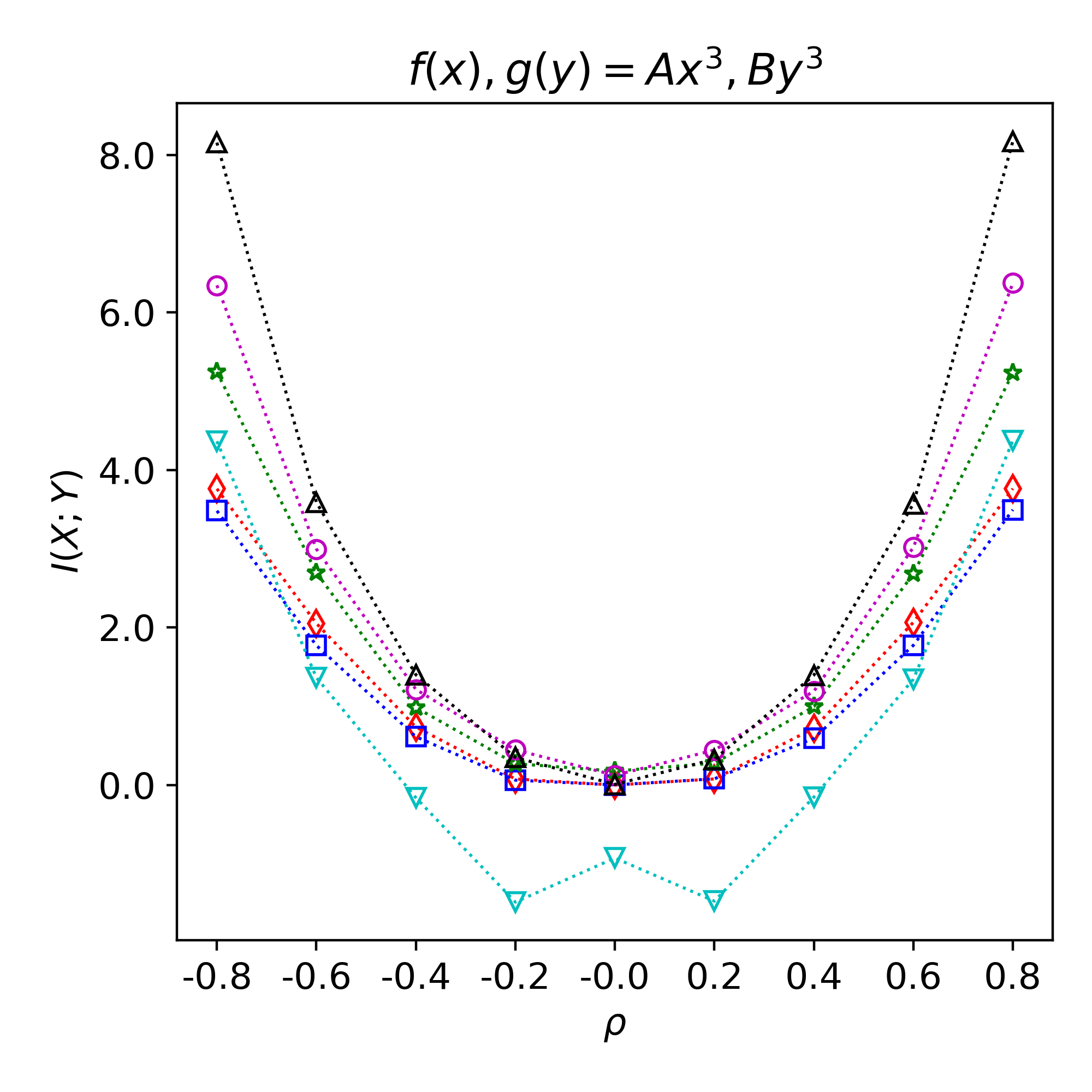}
\caption{\centering $\textbf{A}x^3, \textbf{B}y^3$ }
\end{subfigure}
\hspace{-0.015\textwidth}
\begin{subfigure}{.255\textwidth}
\centering
\includegraphics[width=1.0\linewidth]{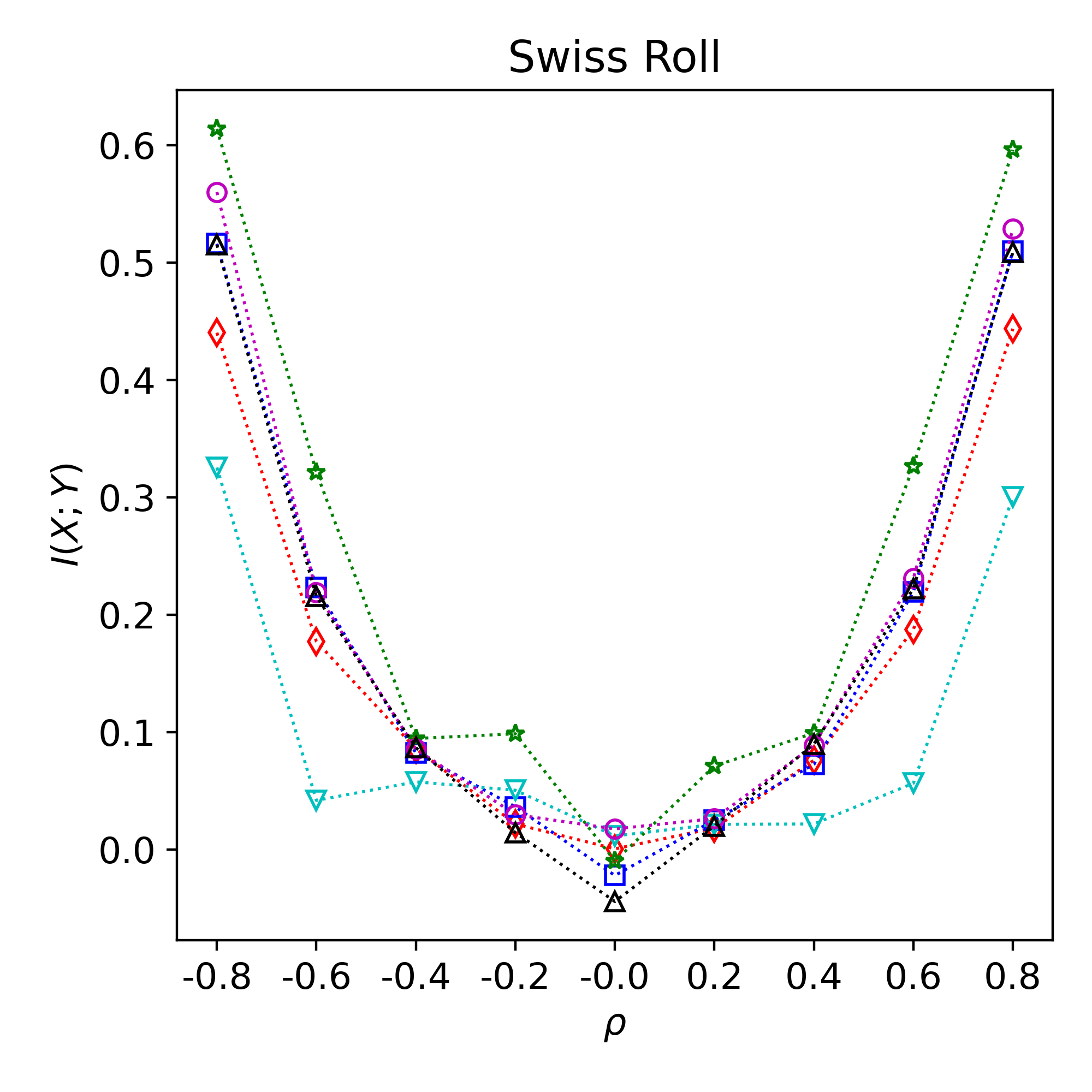}
\caption{\centering Swiss Roll}
\end{subfigure}
\caption{Comparison of different MI estimators under different $\rho$ in four representative synthetic datasets. The dimensionality $d$ of the data $X, Y \in \R^d$ in the four cases are 64, 32, 32, 2 respectively. }
\label{fig:synthetic:rho}
\end{figure}

\begin{figure}[t!]
\centering
\hspace{-0.03\textwidth}
\begin{subfigure}{.255\textwidth}
\centering
\includegraphics[width=1.0\linewidth]{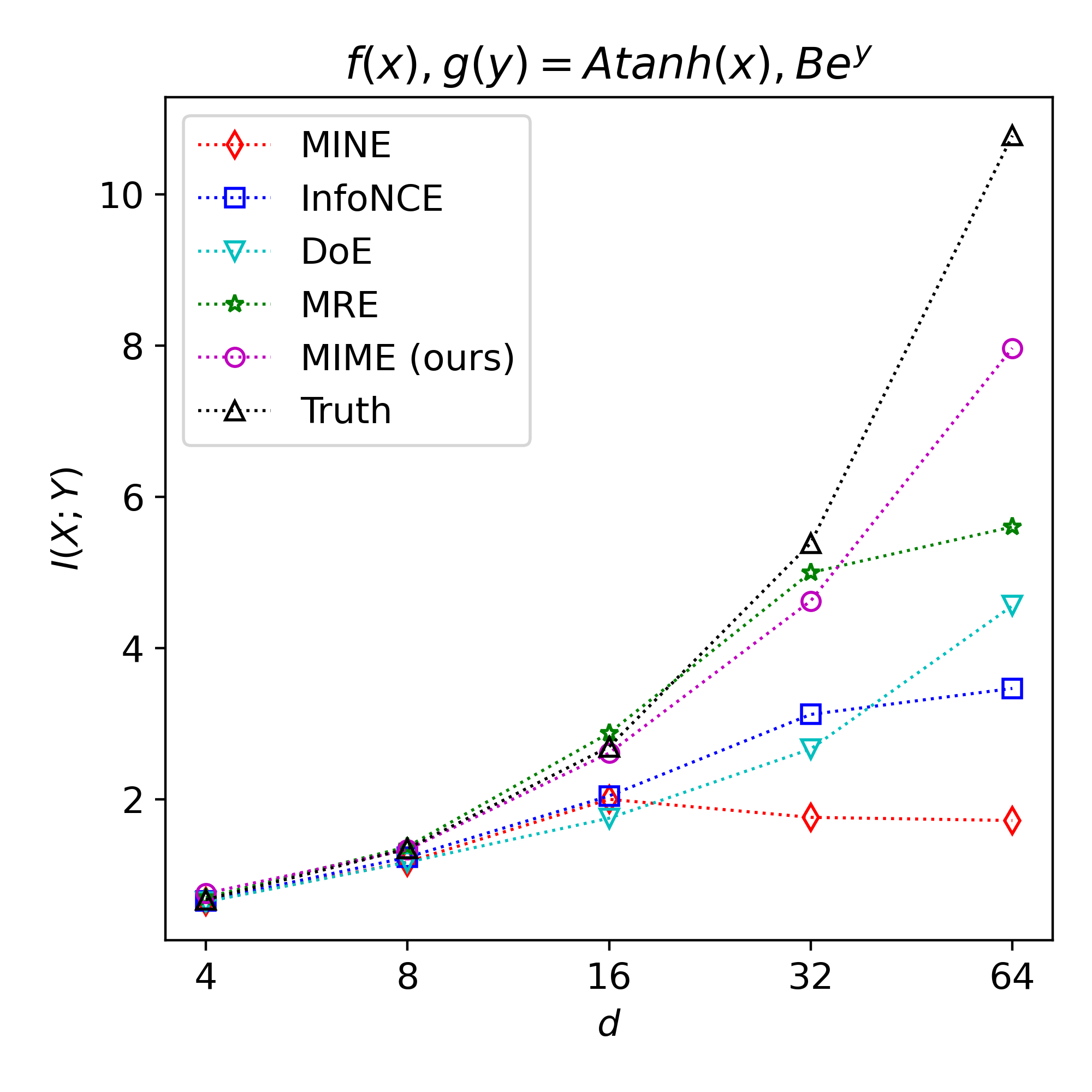}
\caption{\centering $\textbf{A}\text{tanh}(x), \textbf{B}e^y$}
\end{subfigure}
\hspace{-0.015\textwidth}
\begin{subfigure}{.255\textwidth}
\centering
\includegraphics[width=1.0\linewidth]{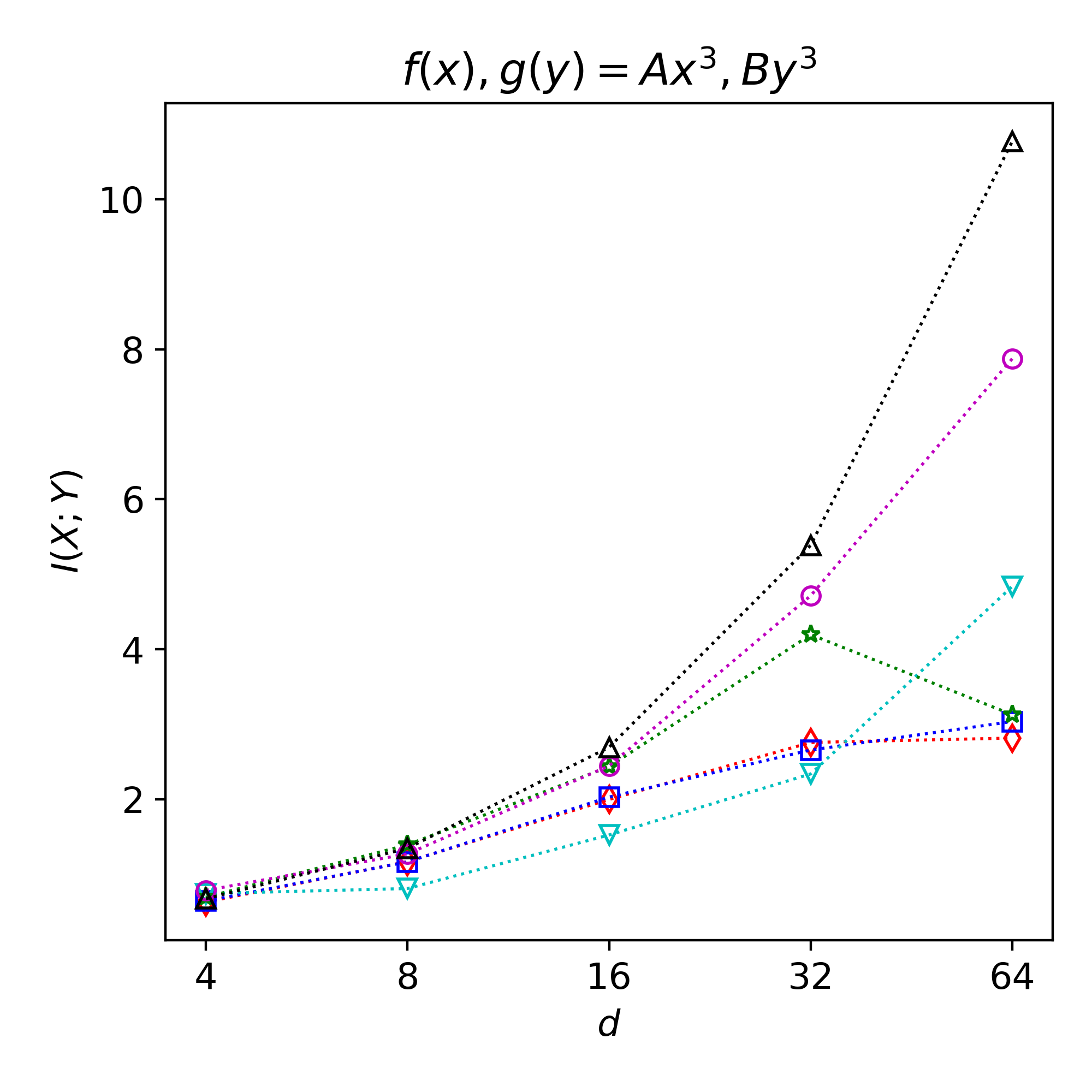}
\caption{\centering $\textbf{A}x^3, \textbf{B}y^3$}
\end{subfigure}
\hspace{-0.015\textwidth}
\begin{subfigure}{.255\textwidth}
\centering
\includegraphics[width=1.0\linewidth]{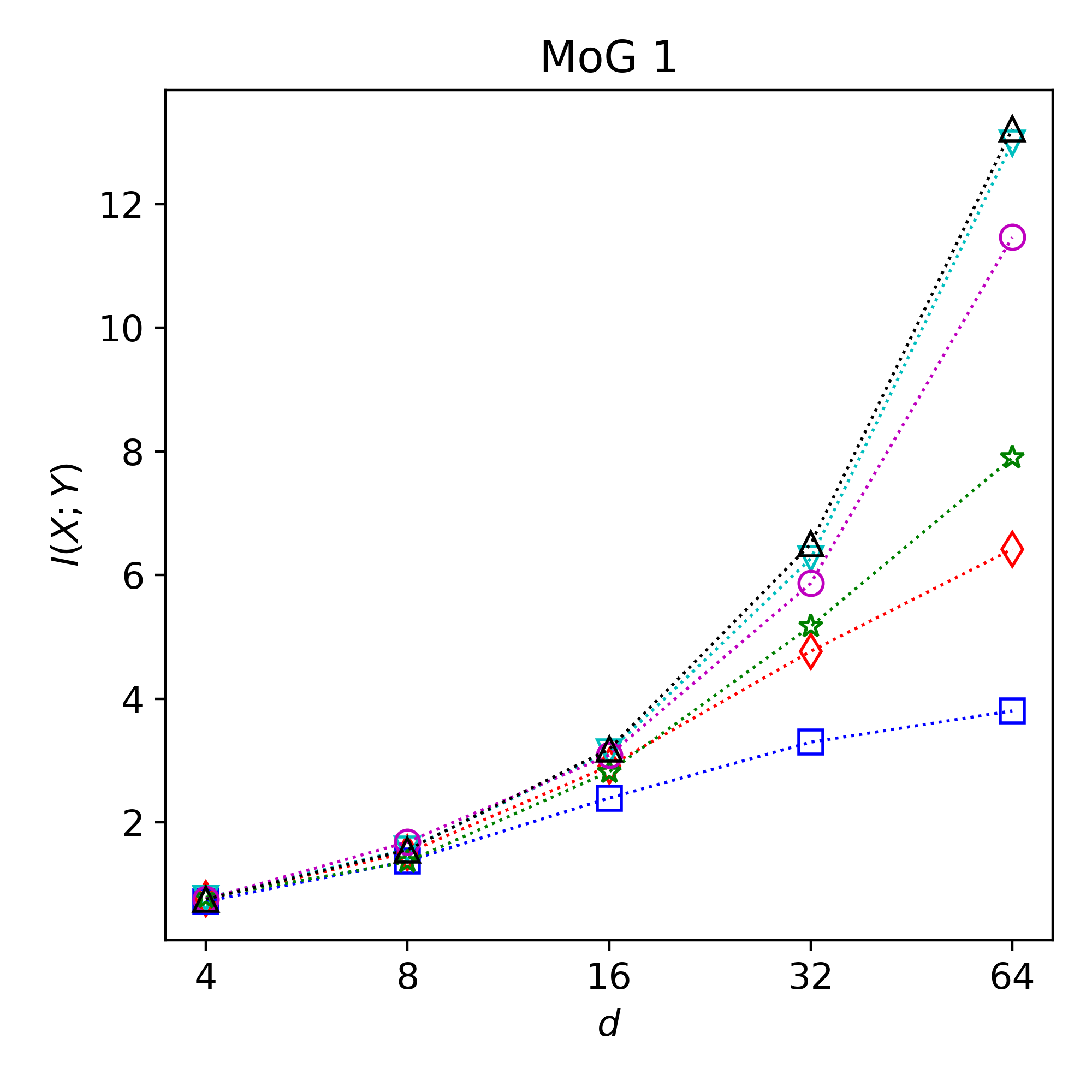}
\caption{\centering  MoG 1}
\end{subfigure}
\hspace{-0.015\textwidth}
\begin{subfigure}{.255\textwidth}
\centering
\includegraphics[width=1.0\linewidth]{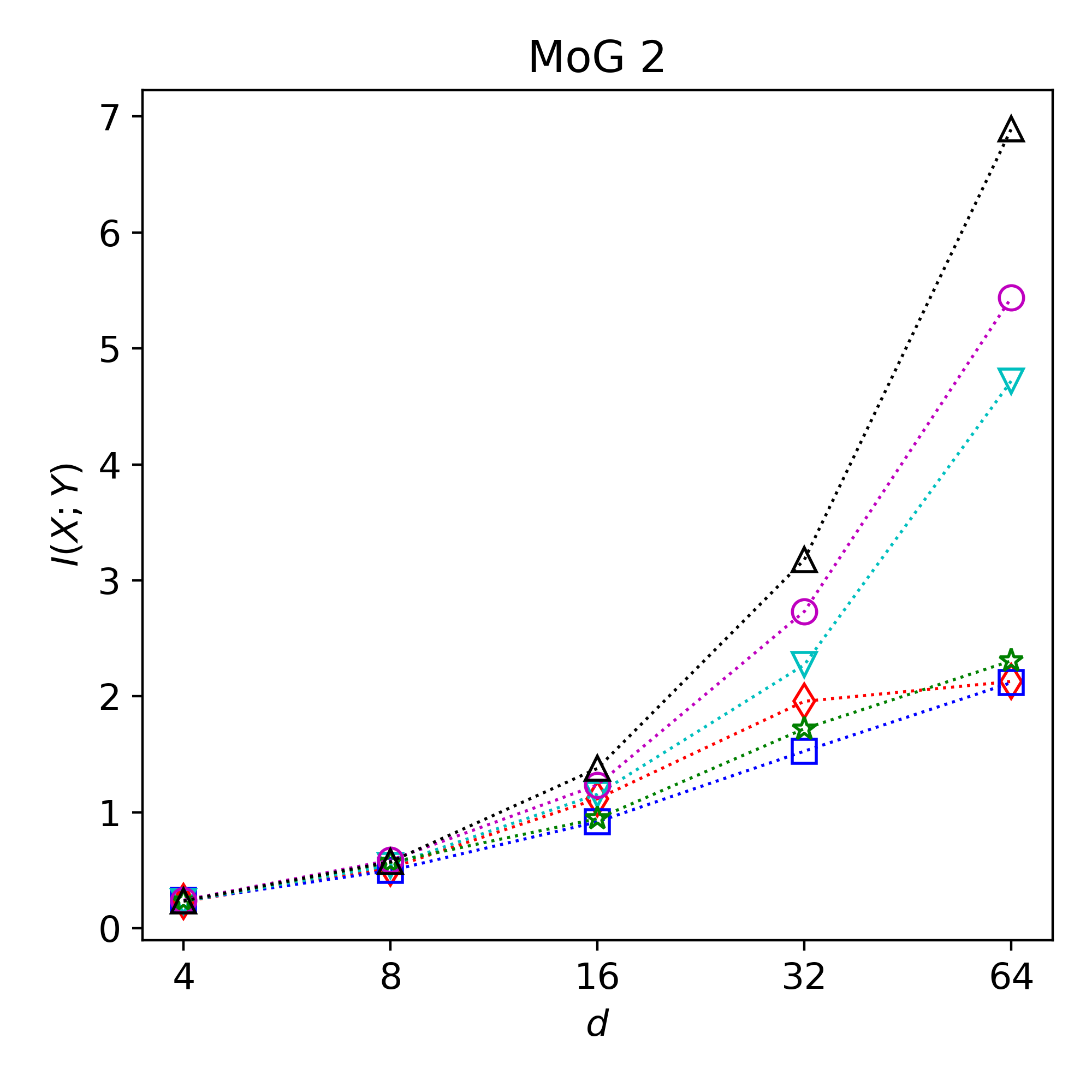}
\caption{\centering MoG 2}
\end{subfigure}
\caption{Comparison of different MI estimators under $\rho = 0.7$ and various data dimensionality $d$.  }
\label{fig:synthetic:dim}
\end{figure}

These benchmarks are similar to those in state-of-the-art benchmarks \cite{czyz2023beyond, czyz2023mixtures} and can be seen as an instantiation of the so-called fine distributions \cite{czyz2023mixtures} for principally evaluating MI estimators. By choosing different functions $f, g$ in the non-linear Gaussian model and different mixture components, we can generate diverse benchmarks with controllable  dependence and complexity easily.

We generate a total number of $n=10^4$ data from each model and estimate MI from the generated data. We are interested in investigating whether our estimator consistently yields a more accurate estimate of MI when compared to other methods, as well as identifying any potential failure mode. All experiment results are collected from 10 independent runs and we report their average values.

\textbf{Varying the dependence level $\rho$}. We first investigate whether our method can  estimate MI accurately when we vary the dependence level $\rho$. Figure \ref{fig:synthetic:rho} shows the results for the non-linear Gaussian task with four typical choices of non-linear transformations $f, g$. These transformations are carefully chosen to represent a wide range of data patterns: Gaussian (case a), data with both bounded and exponential values (case b), data with long tails (case c) and data confined to a low-dimensional manifold (case d); see Appendix B for more details. Estimators that plot closer to the ground truth MI curve are deemed more accurate. Overall, we see that there is no single method to rule in all cases. For example, while DoE performs well with Gaussian data (case a), it shows limitations in non-Gaussian cases (case b,c and d). Similarly, the InfoNCE method is effective when the data is low-dimensional (case d), yet it performs poorly in  higher dimensional scenarios (case a,b and c). Unlike these methods, the proposed estimator consistently ranks within the top two methods across all scenarios, positioning it as a robust method that is widely applicable. The suboptimal performance of our method in the Swiss Roll model may be due to the mismatch between the support of the vector Gaussian copula used and the support of data, where the latter is on a Riemann manifold.

\textbf{Varying the dimensionality $d$}. In Figures \ref{fig:synthetic:dim}, we further compare the performance of the various MI estimators under different data dimensionality. In this task, a higher $d$ leads to a higher MI and thereby a more challenging task of MI estimation. From the figure, we see that the proposed MIME method clearly scales better w.r.t $d$ than most methods, suggesting that it can better estimate MI in high MI settings. Although our method loses to DoE in one of the MoG task (case d), the margin is indeed small, and the MoG method performs unsatisfactorily in case a and case b. The poor performance of DoE may be due to the deficiency of the generative models used (i.e. the neural density estimator), that they can not accurately learn high-dimensional distributions in some cases. On the other hand, we also see that classic distribution-free methods like MINE and InfoNCE can not work reliably when the true MI is beyond 4nats, possibly due to the aforementioned high-discrepancy issue. Our method combines the strengths of both generative methods and distribution-free methods by using generative models as only the reference distributions in distribution-free methods, exhibiting as a strong method that are consistently reliable across diverse cases. We note that even with our powerful method, estimating MI in high-dimensional cases (e.g. $d=64$) still presents a challenge.

\subsection{Bayesian experimental design}
\label{section:bed}
We next consider the application of our method in Bayesian experimental design (BED), an important task in science and engineering. In BED, we wish to design experiments so that the data collected in the experiment will be maximally informative for understanding the underlying process of the experiment. For example, we may wish to find out the best policy design $\mathbf{d}$ for epidemic disease surveillance and intervention, so that we can infer the properties $\theta$ of the epidemic as accurate as possible with minimal observed data $X$. Such a task can be mathematically described as optimizing the conditional mutual information $I(X; \theta|\mathbf{d})$ between the experiment outcomes $X$ (e.g. the observed dynamics of the infectious people in the population) and the properties of interest $\theta$ (e.g. the infectious rate of the epidemic) w.r.t the experiment design $\mathbf{d}$ (e.g. the intervention policy):
\[
    \max_{\mathbf{d}} I(X; \theta|\mathbf{d})
\]
\textbf{Setup}. Many methods have been developed for finding the optimal design in BED \cite{foster2019variational, foster2021deep, ivanova2021implicit, kleinegesse2020bayesian, chaloner1995bayesian, valentin2024designing}. Here, we focus on the most general setup that the data generating process in the experiment is a black-box i.e. we know nothing about the details of the underlying data generation process but only the outcomes and the experiment design used. To find the optimal design $\vecd^*$ in such case, we use Bayesian optimization (BO) 
\cite{srinivas2012information, snoek2012practical} as in existing works to optimize $I(X; \theta|\vecd)$ w.r.t the design $\vecd$. In this work, we use a total number of $T=30$ BO steps to find the optimal design $\vecd^*$, where in each step we run $n=3000$ experiments under design $\vecd$ to collect data $\mathcal{D} = \{x^{(i)}, \theta^{(i)}\}^n_{i=1}$ from the experiment. This data is then used to estimate $I(X; \theta|\mathbf{d})$ by a specific mutual information estimator. 

We consider BED in two popular models that are widely used in epidemiology and medicine studies:
\begin{itemize}[leftmargin=*]
    \item \emph{Death process model}. In this experiment, we model the dynamics of an epidemic in which healthy individuals become infected at rate $\theta$. The design problem here is to choose observations times $\vecd \in \R^{2}$ at which to observe the number of infected individuals in two consecutive experiments, where in each experiment we observe the infection dynamics $X$ of three independent populations. 
    \item \emph{Pharmacokinetic model}. In this experiment, we consider finding the optimal blood sampling time $\vecd \in \R^{M}, 0 \leq d_i \leq 24$ for a group of patients during a pharmacokinetic (PK) study. Each $d_i$ corresponds to the sample time of each patient in the group. We would like to infer the properties of a specific drug demonstrated to the patients through the blood measurements. Here $M=12$. 
\end{itemize}

We are interested in seeing how the use of different MI estimators will affect the utility $\hat{I}(X; \vectheta|\vecd^*)$ of the found  design $\vecd^*$ as measured by MI. As the comparison on $\hat{I}(X; \vectheta|\vecd^*)$ is only sensible when all methods estimate a lower bound of MI, we use the Donsker-Varadhan  representation \eqref{eq:dv} rather than \eqref{eq:mi-est-ours} to estimate MI in our method, where we set $f(x, y)$ in \eqref{eq:dv}  as the log density ratio estimated by our method. This ensures that the  MI estimated by our method is always a lower bound estimate.

\begin{figure}[t]
\centering
\hspace{-0.045\textwidth}
\begin{subfigure}{.31\textwidth}
\centering
\includegraphics[width=1.0\linewidth]{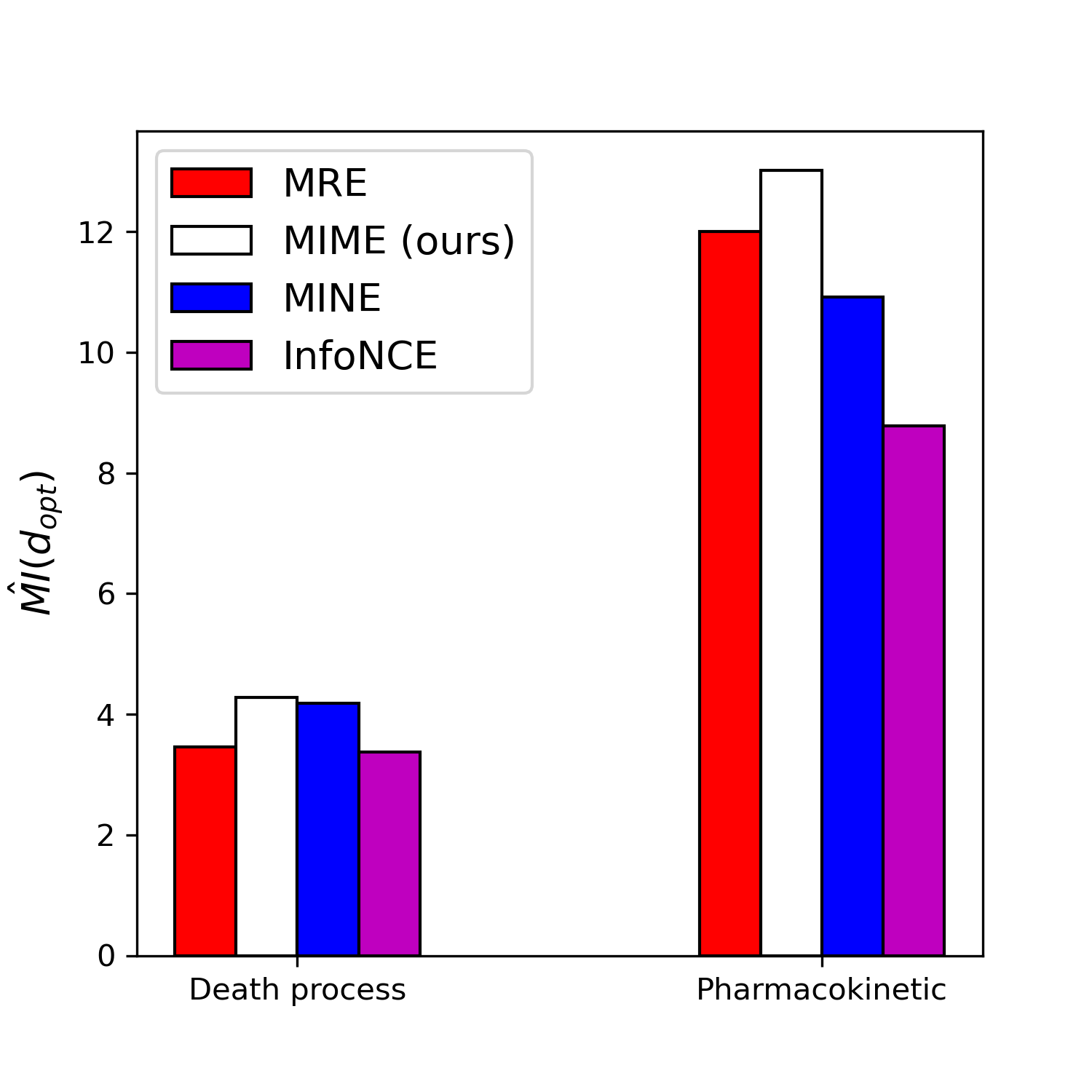}
\caption{\centering}
\end{subfigure}
\hspace{0.000\textwidth}
\begin{subfigure}{.31\textwidth}
\centering
\includegraphics[width=0.920\linewidth]{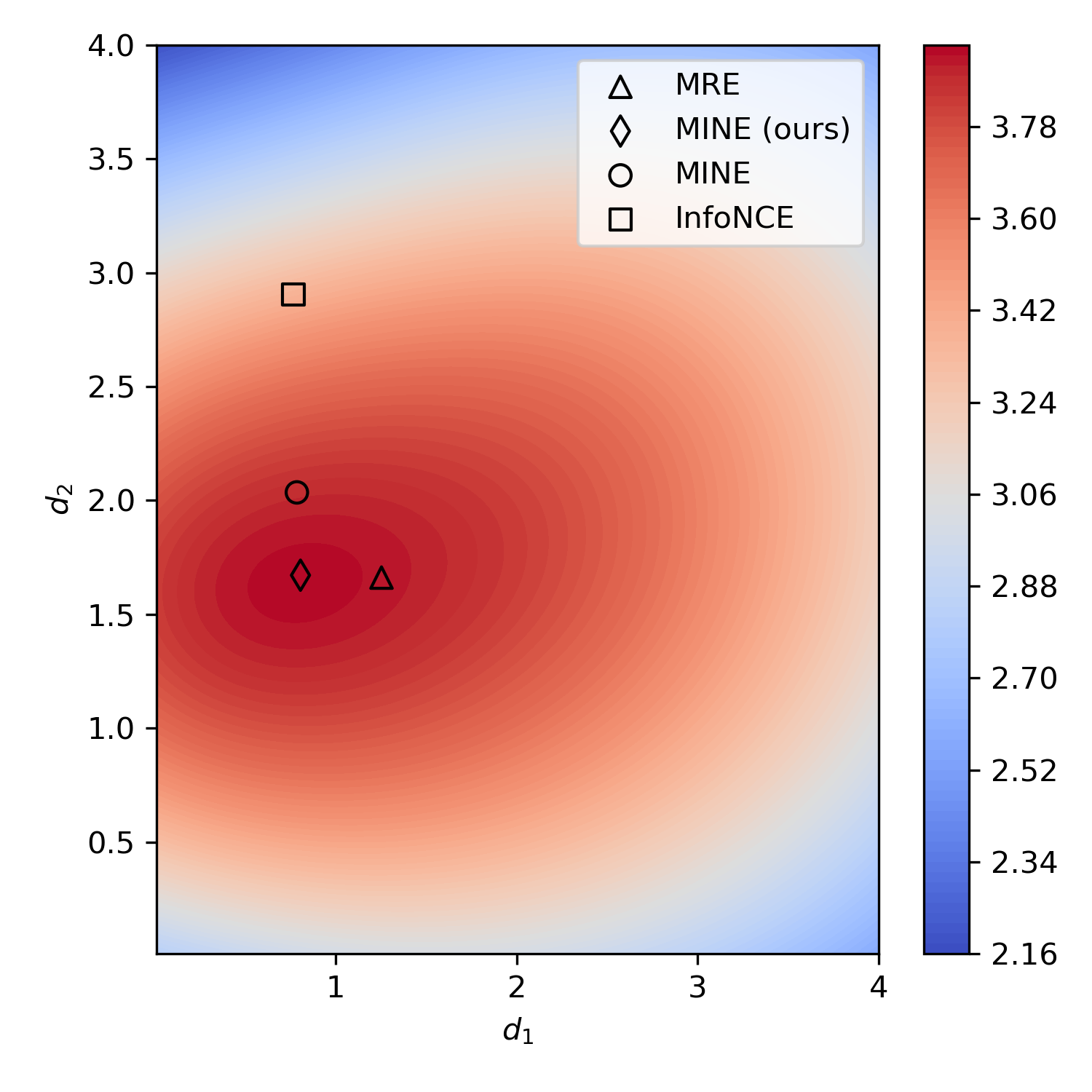}
\caption{\centering }
\end{subfigure}
\hspace{0.0\textwidth}
\begin{subfigure}{.31\textwidth}
\centering
\vspace{0.015\textwidth}
\includegraphics[width=1.0\linewidth]{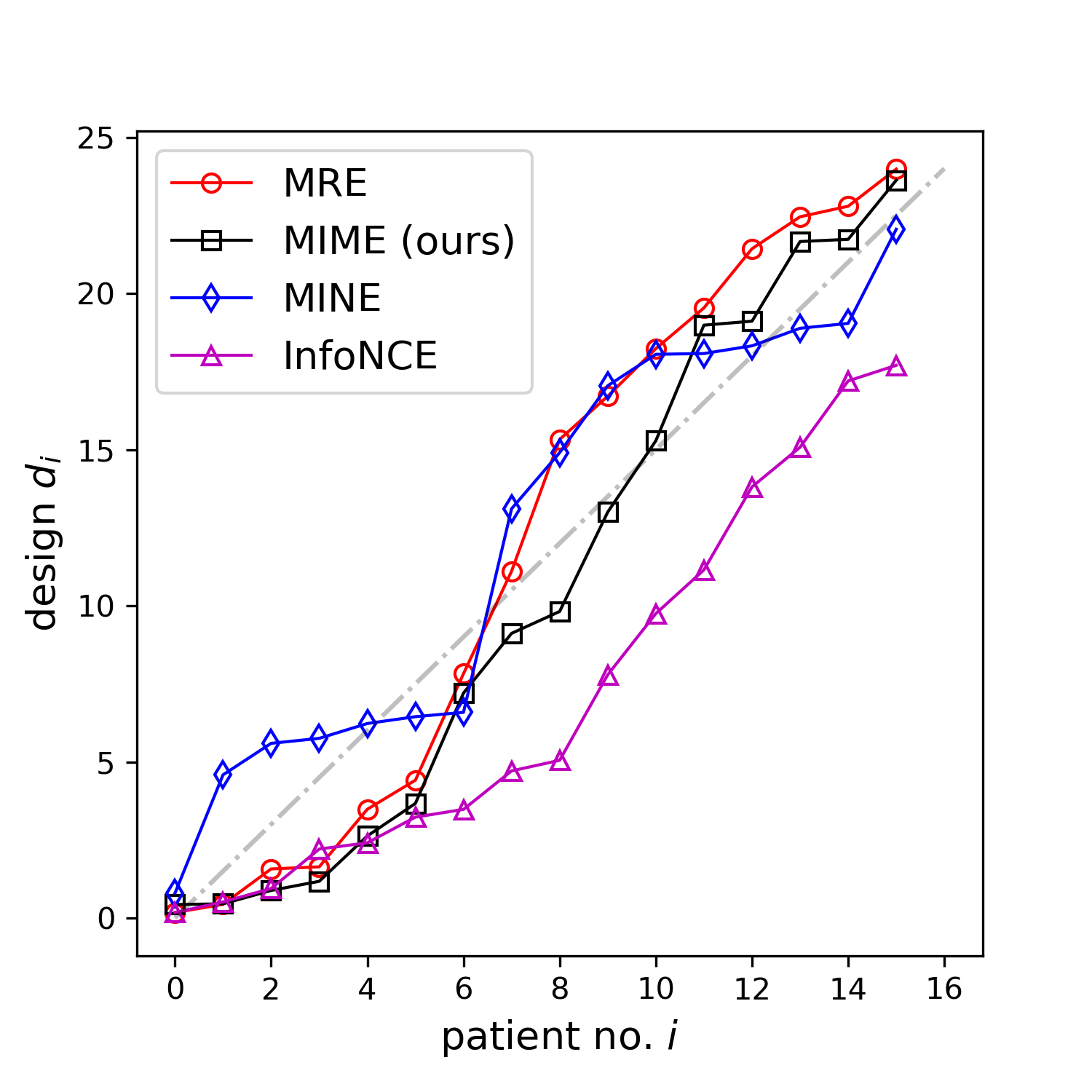}
\caption{\centering  }
\end{subfigure}
\caption{Experiment results for BED. (a) Comparing the utility of the optimal design $\vecd^*$ found by different MI estimators. (b) Visualizing the contour of the underlying function between the utility and the design $\vecd$ in the death process task. (c) Visualizing the optimal designs found in the PK task.}
\label{fig:bed}
\end{figure}

\textbf{Results}. From Figure \ref{fig:bed}.(a), it is evident that our method consistently finds a better design $\vecd^*$ for both models, especially for the PK model where the underlying MI is high. In Figure \ref{fig:bed}.(b), we further visualize the underlying function between the utility and the design in the death process model. This function is modelled by a GP trained with 300 data. As one can see, the designs found by other methods are indeed suboptimal, being away from the optimum. In Figure \ref{fig:bed}.(c), we manually inspect the design strategies reported by different methods in the PK model. Compared to other methods, the blood sampling times suggested by our method is more uniformly distributed, which is sensible as a more diverse blood sampling time will allow us to collect richer information about the drug.

\subsection{Self-supervised learning}
\label{section:ssl}
The last task we consider is the application of our MI estimator in self-supervised representation learning (SSL). SSL has been shown to be closely connected to infomax learning~\cite{tschannen2019mutual, kong2019mutual, wu2020mutual, hjelm2018learning,chen2020simple,shidanipoly}. For instance, \cite{hjelm2018learning,chen2020simple} showcase that maximizing the mutual information between different ``views" of an image enables the acquisition of meaningful representations for downstream tasks. In this experiment, we aim to re-analyze SSL with our empowered estimator, investigating (a) how good can existing methods in SSL e.g. SimCLR \cite{chen2020simple} maximize the MI between different views of data; and (b) whether higher mutual information consistently results in better representations in SSL.

\textbf{Setup}. We focus on analyzing SimCLR, a classic method for SSL \cite{chen2020simple}. SimCLR learns representation by maximising the MI among various views of the same image by the InfoNCE loss~\cite{chen2020simple, van2018representation}
\[
    \max_{f, h} I_{\text{InfoNCE}}(h(f(v_1)), h(f(v_2))) 
\]
where $f$ is an encoder (implemented as a ResNet-50  model \cite{he2016deep}) used to compute the representation of an image. $h$ is some projection function (often known as the projection head). Here the dimensionality of the representation is set to be $D=1024$ and the batch size is set to be $B=256$. 

In addition to standard SimCLR where the estimate of MI is limited by the log batch size \cite{poole2019variational}, we also consider a variant of SimCLR, SimCLR+, which can potentially attain a higher MI:
\[
    \max_{f, h} \beta \cdot I_{\text{MIME}}(f(v_1), f(v_2)) + (1-\beta) \cdot I_{\text{InfoNCE}}(h(f(v_1)), h(f(v_2)))
\]
Here, we use the DV representation \eqref{eq:dv} to estimate MI in our method, where we  set $f(x, y)$ in \eqref{eq:dv}  as the log density ratio \eqref{eq:ratio} estimated in our method. This ensures that our estimate is a lower bound. We adopt the common linear evaluation protocol to assess the quality of the learned representation.

\begin{figure}[t]
\centering
\hspace{-0.025\textwidth}
\begin{subfigure}{.32\textwidth}
\centering
\includegraphics[width=1.0\linewidth]{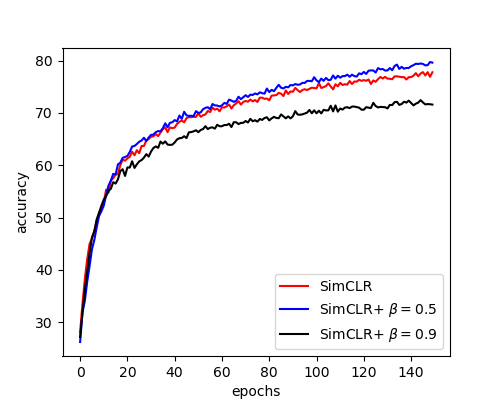}
\caption{\centering}
\end{subfigure}
\hspace{-0.015\textwidth}
\begin{subfigure}{.32\textwidth}
\centering
\includegraphics[width=1.0\linewidth]{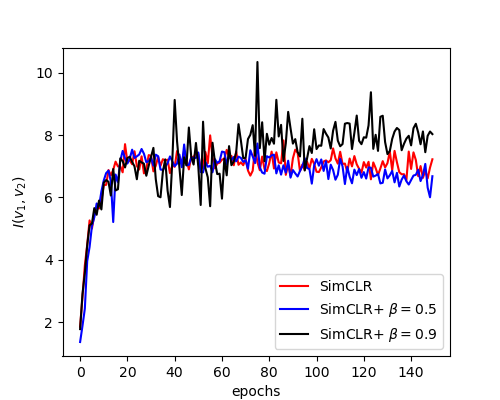}
\caption{\centering }
\end{subfigure}
\hspace{0.0\textwidth}
\begin{subfigure}{.32\textwidth}
\centering
\vspace{-0.015\textwidth}
\includegraphics[width=1.0\linewidth]{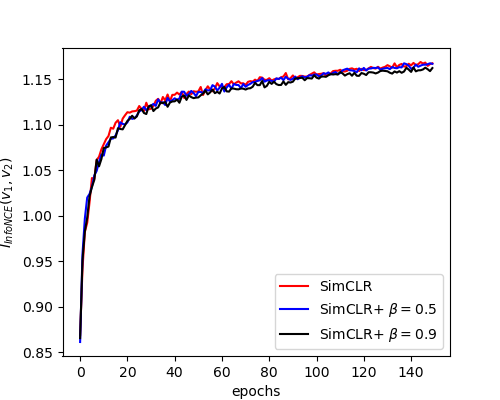}
\caption{\centering  }
\end{subfigure}
\caption{Experimental results for SSL. (a) Classification accuracy on test set as evaluated by a linear classifier. (b) $I(v_1, v_2)$ as estimated by our method. 
(c) $I(v_1, v_2)$ as computed by InfoNCE.}
\label{fig:SSL}
\end{figure}

\textbf{High MI does not imply good representation}. In Figure \ref{fig:SSL}, we respectively plot (a)  the test set accuracy; (b) the mutual information (MI) between different views of data as estimated by our method; (c) the MI estimated by InfoNCE. These figures are collected from a typical run on CIFAR10. By comparing SimCLR and SimCLR+ with $\beta=0.9$, we see that a high MI may not necessarily be a good sign of better representation: SimCLR+ clearly attains a higher MI but has a significantly lower test accuracy. The weak correlation between high MI and representation quality is also evidenced by the trends of the curves in Figure \ref{fig:SSL}.(a) and Figure \ref{fig:SSL}.(b), where the test accuracy continues to grow even when there is no increase on MI. The results echo with previous studies \cite{tschannen2019mutual} which stated that the success of SSL may only loosely relate to MI maximization under the linear evaluation protocol.
 
\textbf{InfoNCE is useful for MI maximization despite underestimation}. Another interesting result is seen by comparing Figure \ref{fig:SSL}.(b) and Figure \ref{fig:SSL}.(c), where we find that InfoNCE actually largely underestimates the true MI between different  views of data\footnote{We have included the constant $\log B$ when we compute the MI in InfoNCE (here $B$ is the batch size).}, with the true MI being several times higher. Interestingly, despite InfoNCE largely underestimates MI, optimizing InfoNCE loss does lead to an effective maximization of the true MI at the early stage of learning. Furthermore, the learning dynamics of InfoNCE is also more stable than that of SimCLR+ which detects high MI. This suggests InfoNCE loss is a good proxy for optimizing MI despite it largely underestimates MI.

\section{Conclusion}
In this paper, we propose a new method for mutual information (MI) estimation in high-MI settings.
A main challenge in this task is the high-discrepancy issue, which causes direct comparison between the joint distribution and the marginal distribution inaccurate. We address this issue by additionally comparing these distributions with their vector copula approximations, which share the same marginal distributions as the original data distribution and have Gaussian dependence structures. Experiments on diverse tasks demonstrate that such additional comparisons provide fine-grained  signals for accurately modelling the true dependence in data, which  leads to accurate MI estimate.

Despite our powerful method, we note that MI estimate is still challenging with high-dimensional non-Gaussian data, as can be seen from the simulation studies in Section \ref{section:synthetic}. Furthermore, we find that to maximize MI, an accurate estimate of its value may indeed be unnecessary --- a lesson learned from the SSL task. Together, these results highlight the values of developing proxies of MI for information-theoretic representation learning such as sliced mutual information~\cite{goldfeld2021sliced, chen2022scalable, chen2023learning, tsur2024max}.

While we believe our method can be applied in many domains for societal benefit (e.g. in designing medical experiments), since it is a general technique, one must be careful to prevent potential harms.

\clearpage

\bibliography{main}
\bibliographystyle{unsrt}

\clearpage
\appendix

\begin{center}
\LARGE
\textbf{Supplementary Materials}
\end{center}

\vspace{0.25cm}

\section{Theoretical proofs}

In this section, we are going to prove Proposition 1 and Proposition 2. Before delving into that, we first revisit the proposed MIME estimator. Specifically, MIME proposes to estimate the mutual information by learning a multinomial classifier $h$ with a set of delicate reference distributions. To train the classifier $h$, we approximate the learning objective by using $N$ samples 
\begin{align}
    \hat{h} = \argmax_h \mathcal{L}_N (h) := \frac{1}{N} \sum_{i=1}^N \log \frac{e^{h_c (x_i, y_i)}}{\sum_{k=1}^K e^{h_k (x_i, y_i)}}, \quad x_i, y_i \sim p(c)p_c(x,y), \nonumber
\end{align}
where we use a four-ways classifier (i.e., $K=4$) with $p_1(x,y) = p(x,y)$ and $p_4(x,y)=p(x)p(y)$. After training, we approximate the true mutual information $I(X;Y)$ using 
\begin{align}
    I(X;Y) \approx \hat{I}(X;Y, \hat{h}) := \frac{1}{N} \sum_{i=1}^N \hat{h}_1 (x_i, y_i) - \hat{h}_4 (x_i, y_i). \nonumber
\end{align}
For brevity, we denote $\hat{r}(x,y) := \hat{h}_1(x,y) - \hat{h}_4(x,y)$ as the estimated ratio with the learned classifier $\hat{h}$ and ${r}^* (x,y) := \log \frac{p(x,y)}{p(x)p(y)}$ as the ground truth ratio.
Now, we are ready to prove the consistency of the proposed MIME estimator.

\subsection{Proof of Proposition 1}
\begin{Proposition}
(\emph{Consistency of multinomial MI estimate}). Assuming that the multinomial classifier $h_c: X \times Y \rightarrow \mathbb{R}$ is uniformly bounded. Then, for every  $\varepsilon > 0$ there exists $N(\varepsilon) \in \mathbb N$, such that 
\[
    \left|\hat{I}_{N}(X; Y|\hat{h}) - I(X; Y)\right| < \varepsilon, \forall N \geq N(\varepsilon), a.s..
\]
\end{Proposition}
\emph{Proof}. Firstly, we apply the triangle inequality to upper bound the difference $|\hat{I}_{N}(X; Y|\hat{h}) - I(X; Y)|$ as follows
\begin{align}
    \left|\hat{I}_{N}(X; Y|\hat{h}) - I(X; Y)\right| \leq \left| \hat{I}_{N}(X; Y|\hat{h}) - \frac{1}{N} \sum_{i=1}^N \hat{r}(x_i, y_i) \right| + \left| \frac{1}{N} \sum_{i=1}^N \hat{r}(x_i, y_i) - I(X;Y) \right|. \nonumber
\end{align}
Obviously, the first term in the RHS can be bounded by a sequence $\varepsilon_N \xrightarrow{a.s.} 0$ due to the normal strong law of large numbers. To bound the second term, we employ the result in \cite[Appendix A]{srivastava2023estimating}, which proves that the multinomial ratio estimator is consistent. Specifically, let $\hat{h} = \argmax_h \mathcal{L}_N (h)$ and $\hat{r} = \hat{h}_1 - \hat{h}_4$. Then for every $ \varepsilon > 0$, there exists $N \in \mathbb{N}$, such that
\begin{align}
    \left| \hat{r}(x, y) - r^* (x, y) \right| < \varepsilon, \forall x,y, a.s.. \nonumber
\end{align}
Thereby, there exists a sequence $\varepsilon_i \xrightarrow{a.s.} 0$ for each $i=1,2,\dots,N$ such that
\begin{align}
    \left|\frac{1}{N} \sum_{i=1}^N \hat{r}(x_i, y_i) - I(X;Y)\right| 
    &= \left|\frac{1}{N} \sum_{i=1}^N \hat{r}(x_i, y_i) - \mathbb{E}_{p(x,y)} \left[ \log \frac{p(x,y)}{p(x)p(y)} \right] \right|  \nonumber \\
    &= \left|\frac{1}{N} \sum_{i=1}^N \hat{r}(x_i, y_i) - \frac{1}{N} \sum_{i=1}^N r^*(x_i, y_i) \right| \nonumber \\
    &\leq \frac{1}{N} \sum_{i=1}^N \left| \hat{r}(x_i, y_i) - r^*(x,y) \right| \nonumber \\
    &\leq \frac{1}{N} \sum_{i=1}^N \varepsilon_i \leq \max (\varepsilon_1, \dots, \varepsilon_N), \nonumber
\end{align}
where from line 1 to line 2, we utilise the law of large numbers, and from line 3 to line 4, the consistency of multinomial ratio estimator is employed. Hence, for each $\epsilon > 0$, there exist an $N(\varepsilon) \in \mathbb{N}$, such that $\left|\hat{I}_{N}(X; Y|\hat{h}) - I(X; Y)\right| < \varepsilon$, almost surely.
\qed \\

\subsection{Proof of Proposition 2}
\begin{Proposition}
(\emph{Controlled error in multinomial ratio estimate}). Define $\log \hat{r}_{i, j}(x, y)  \coloneqq h_i(x, y) - h_j(x, y)$ with $h$ being the classifier defined as above. Upon convergence, we have that for any $i$,
\[
    \Big |\log \hat r_{1, 4}(x, y) - \log \frac{p(x, y)}{p(x)p(y)} \Big | \leq \lambda\Big |\log \hat r_{i, i+1}(x, y) - \log \frac{p_{i}(x, y)}{p_{{i}+1}(x, y)} \Big| ,
\]
where $\lambda \leq 3$ is some constant. 
\end{Proposition}
\emph{Proof}. The key of the proof is the following identity in the learned network $\hat{h}$:
\[
    \hat{h}_1 - \hat{h}_4 = \sum^3_{i=1} (\hat{h}_i - \hat{h}_{i+1})  
\]
which, by definition,
\[
    \log \hat{r}_{1, 4}(x, y) = \sum^3_{i=1} \log \hat{r}_{i, i+1}(x, y)
\]
Let $r_{i, j}(x, y) = \frac{p_i(x, y)}{p_j(x, y)}$ be the true density ratio between $p_i(x, y)$ and $p_j(x, y)$. We have the following important observation:
\[
    \log r_{1, 4}(x, y) = \sum^3_{i=1} \log r_{i, i+1}(x, y)
\]
which is due to the fact $\frac{p_1(x,y)}{p_4(x, y)} = \frac{p_1(x,y)}{p_2(x, y)} \cdot \frac{p_2(x,y)}{p_3(x, y)} \cdot \frac{p_3(x,y)}{p_4(x, y)}$.
By triangular inequality
\[
    \Big| \sum^3_{i=1} \log \hat{r}_{i, i+1}(x, y) -  \sum^3_{i=1} \log r_{i, i+1}(x, y)  \Big| \leq \sum^3_{i=1} \Big| \log \hat{r}_{i, i+1}(x, y)-  \log r_{i, i+1}(x, y)  \Big|
\]
which leads to
\[
     \Big |\log \hat r_{1, 4}(x, y) - \log r_{1, 4}(x, y) \Big| \leq 3 \sup_i \Big| \log \hat{r}_{i, i+1}(x, y)-  \log r_{i, i+1}(x, y)  \Big|
\]
substituting $r_1(x, y) = \frac{p(x, y)}{p(x)p(y)}$ and setting $\lambda =3$, we have
\[
    \Big |\log \hat r_{1, 4}(x, y) - \log \frac{p(x, y)}{p(x)p(y)} \Big| \leq \lambda \sup_i \Big| \log \hat{r}_{i, i+1}(x, y)-  \log r_{i, i+1}(x, y)  \Big|
\]
which completes the proof. Note that the constant $\lambda$ is typically smaller than 3 due to the use of triangular inequality above. \qed

\clearpage

\section{Additional experimental details and results}

\subsection{Details of the models in synthetic tasks}
We provide details of the Swiss Roll model, MoG 1 model and MoG 2 model in the synthetic task.

\begin{itemize}[leftmargin=*]
\item \emph{Swiss Roll}. In this model, data $x \in \R^2$, $y \in \R$ is generated by the following data generating process:
\[
    x_1 = \frac{t \cos(t)}{21}, \qquad x_2 = \frac{t \sin(t)}{21}, \qquad  y = v,
\]
where
\[
    t = \frac{3\pi}{2}  (1+2 \Phi(\epsilon_x)), \qquad v = \Phi(\epsilon_y)
\]
\[
     \begin{bmatrix}
\epsilon_x \\
\epsilon_y 
\end{bmatrix}  \sim \mathcal{N}\Big(  \begin{bmatrix}
\epsilon_x \\
\epsilon_y 
\end{bmatrix}, \mathbf{0},   \begin{bmatrix}
1 & \rho \\
\rho & 1
\end{bmatrix} \Big)
\]
Here $\Phi(\cdot)$ is the cumulative distribution function (CDF) of standard normal distribution. It is easy to show that this model is still in the class of non-linear transformation of Gaussian. 

\item \emph{MoG 1}. This model consists of 5 equally-weighted mixture of multivariate Gaussian where the mean for the $k$th component is $\mu_k = m_k\mathbf{1}$ and the dependence coefficient in the $kth$ covariance matrix is $\rho_k$. Here $m_1, ... m_5 = [-0.4, -0.1, 0, 0.1, 0.4]$ and 
$\rho_1, ... \rho_5 = [0.5, 0.6, 0.7, 0.8, 0.9]$.
\item \emph{MoG 2}. This model consists of 5 equally-weighted mixture of multivariate Gaussian where the mean for the $k$th component is $\mu_k = m_k\mathbf{1}$ and the dependence coefficient in the $k$th covariance matrix is $\rho_k$. Here $m_1, ... m_5 =  [-0.2, -0.1, 0, 0.3, 0.4]$ and 
$\rho_1, ... \rho_5 =  [-0.2, -0.1, 0, 0.3, 0.4]$.
\end{itemize}

\def\lc{\left\lceil}   
\def\rc{\right\rceil}

\subsection{Implementation of the mapping $f, g$ in vector Gaussian copula}
The detailed implementation of the bijective mappings $f$ and $g$ are task-dependent. Specifically,
\begin{itemize}[leftmargin=*]
    \item \textbf{Non-representation learning tasks}. For this type of tasks, we implement the two mappings $f, g$ by a continuous flow model, specifically a MAF~\cite{papamakarios2017masked}. This MAF is trained by Adam~\cite{kingma2014adam} with a learning rate of $5\times 10^-4$ and early stopping. We use this implementation for all the experiments in Section 6.1 (synthetic data) and Section 6.2 (Bayesian experiment design).
    \item \textbf{Representation learning tasks}. For this type of tasks, we implement $f, g$ as element-wise functions. In such case, the model degenerates to a Gaussian copula model. Sampling in Gaussian copula can be done by first sampling $\epsilon \sim \mathcal{N}(\epsilon; 0, \Sigma)$, then computing the rank of $\epsilon_d$ in the population: $r_d = \lc n\Phi(\epsilon_d) \rc$. Here $\Phi(\cdot)$ is the CDF of standard normal distribution and $n$ is the sample size of population. After that, we set $x_d$ as the $r_d$-th smallest element in the population of $X_d$. The estimation of $\Sigma$ can be done by inverting this process, which takes $\leq 30$ms when $n=10,000$.   
\end{itemize}

\subsection{Comparisons of different reference distributions}

We conducted experiments with different choices of reference distributions $q(x,y)$. The results are shown in Figure 1. Here MIME (vgc) is our original method where $q(x,y)$ is modeled by a vector Gaussian copula and hence is marginal-preserving. MIME (single flow) is the case where $q(x, y)$ is modelled by a single flow model. This flow is the same as in MIME (vgc). Note that this flow is likely to be more difficult to learn than the two flows $q(x)$ and $q(y)$ in VGC due to the increased dimensionality. MRE is a baseline where $q(x,y)$ is an implicit distribution that is not marginal-preserving. It is clear that MIME (vgc) achieves the best performance.

\subsection{The impact of the number of training data}
Mutual information estimation is sensible to the number of the given samples. To further understand how our method performs in small data regimes, we conducted additional experiments with sample sizes $N = 2,500$ and $N =  1,250$, respectively. As shown in \cref{fig:small_N} and \cref{fig:tiny_N}, the proposed estimator still demonstrates superior performance compared to the baselines in small data regimes.

\quad \\

\subsection{Computation time comparisons}

\begin{table}[h]
\small
\centering
\caption{Running time of a typical run of different MI estimators on the MoG tasks.  }
\label{tab:app-runing-time}
\begin{tabular}{l|ccccc}
\toprule 
Method & InfoNCE & MINE & MRE & DoE & MIME   \\
\midrule 
time (s) & 390.24 & 140.71 & 40.28 & 321.45 & 160.21   \\
\bottomrule
\end{tabular}
\end{table}

\begin{table}[h]
\small
\centering
\caption{Running time of a typical run of different MI estimators on the non-linear Gaussian tasks.  }
\label{tab:app-runing-time-2}
\begin{tabular}{l|ccccc}
\toprule 
Method & InfoNCE & MINE & MRE & DoE & MIME   \\
\midrule 
time (s) & 371.24 & 174.71 & 36.35 & 278.72 & 193.21   \\
\bottomrule
\end{tabular}
\end{table}

\begin{figure}[!t]
\centering
\hspace{-0.03\textwidth}
\begin{subfigure}{.255\textwidth}
\centering
\includegraphics[width=1.0\linewidth]{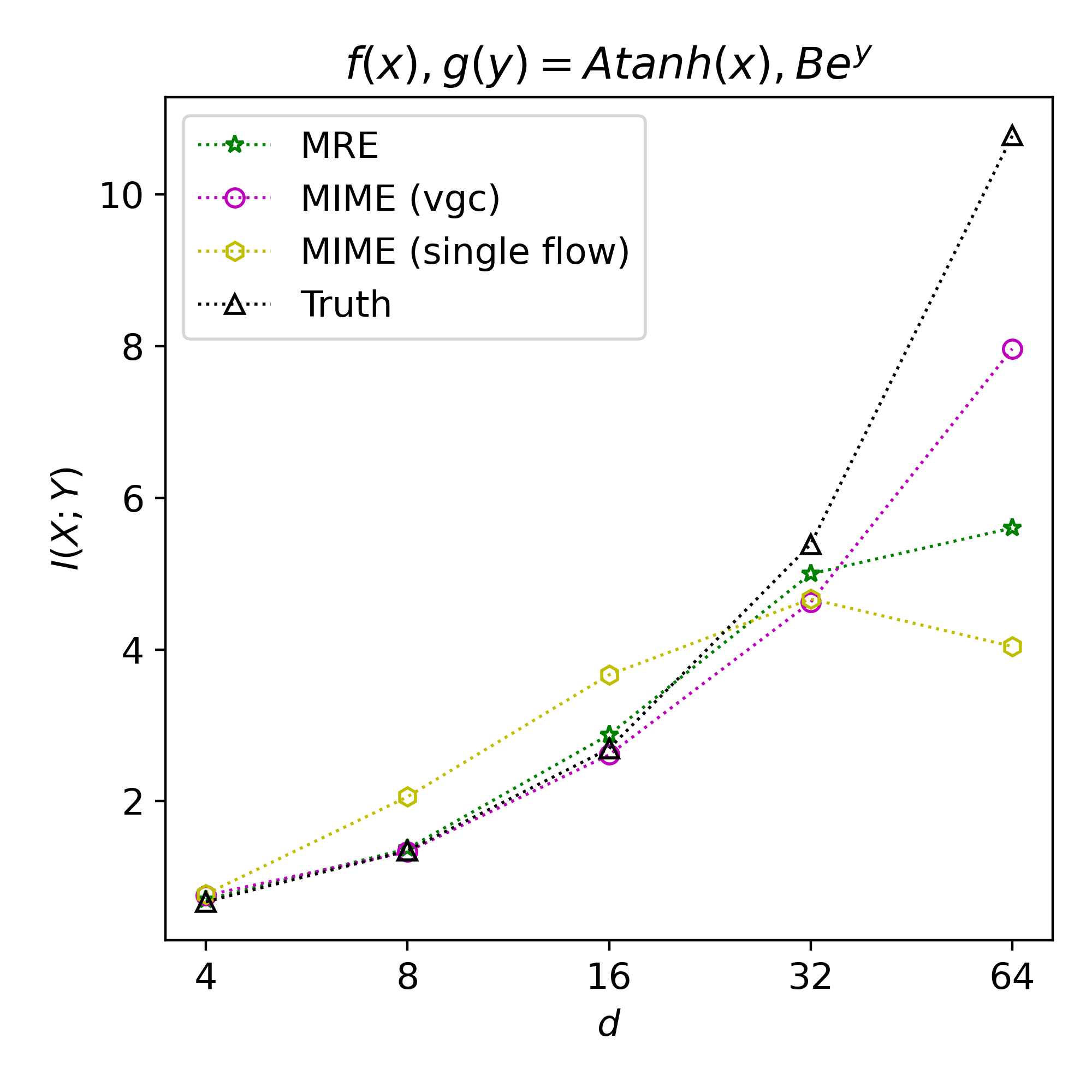}
\vspace{-8mm}
\caption{\centering $A\text{tanh}(x), Be^y$}
\end{subfigure}
\hspace{-0.015\textwidth}
\begin{subfigure}{.255\textwidth}
\centering
\includegraphics[width=1.0\linewidth]{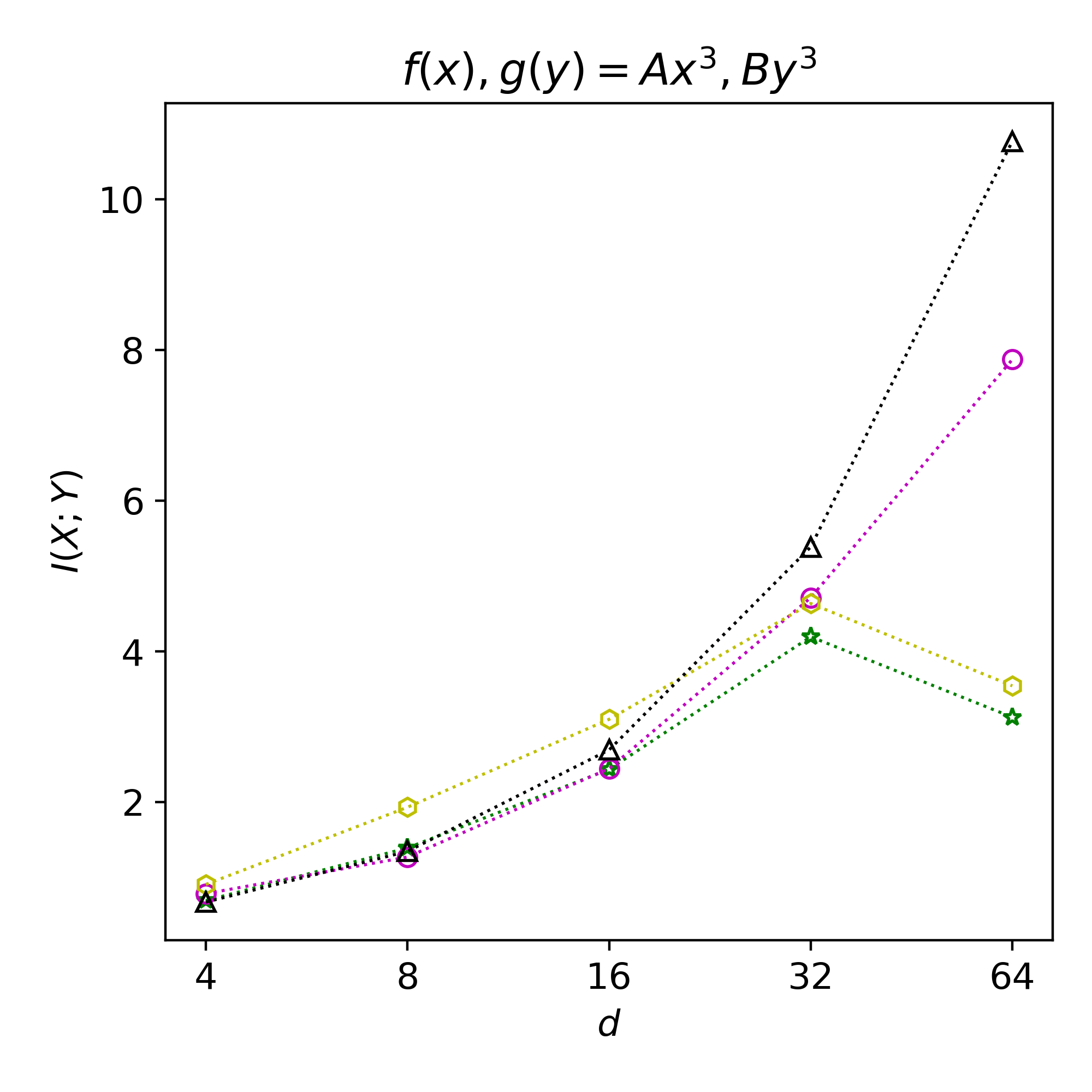}
\vspace{-8mm}
\caption{\centering $Ax^3, By^3$}
\end{subfigure}
\hspace{-0.015\textwidth}
\begin{subfigure}{.255\textwidth}
\centering
\includegraphics[width=1.0\linewidth]{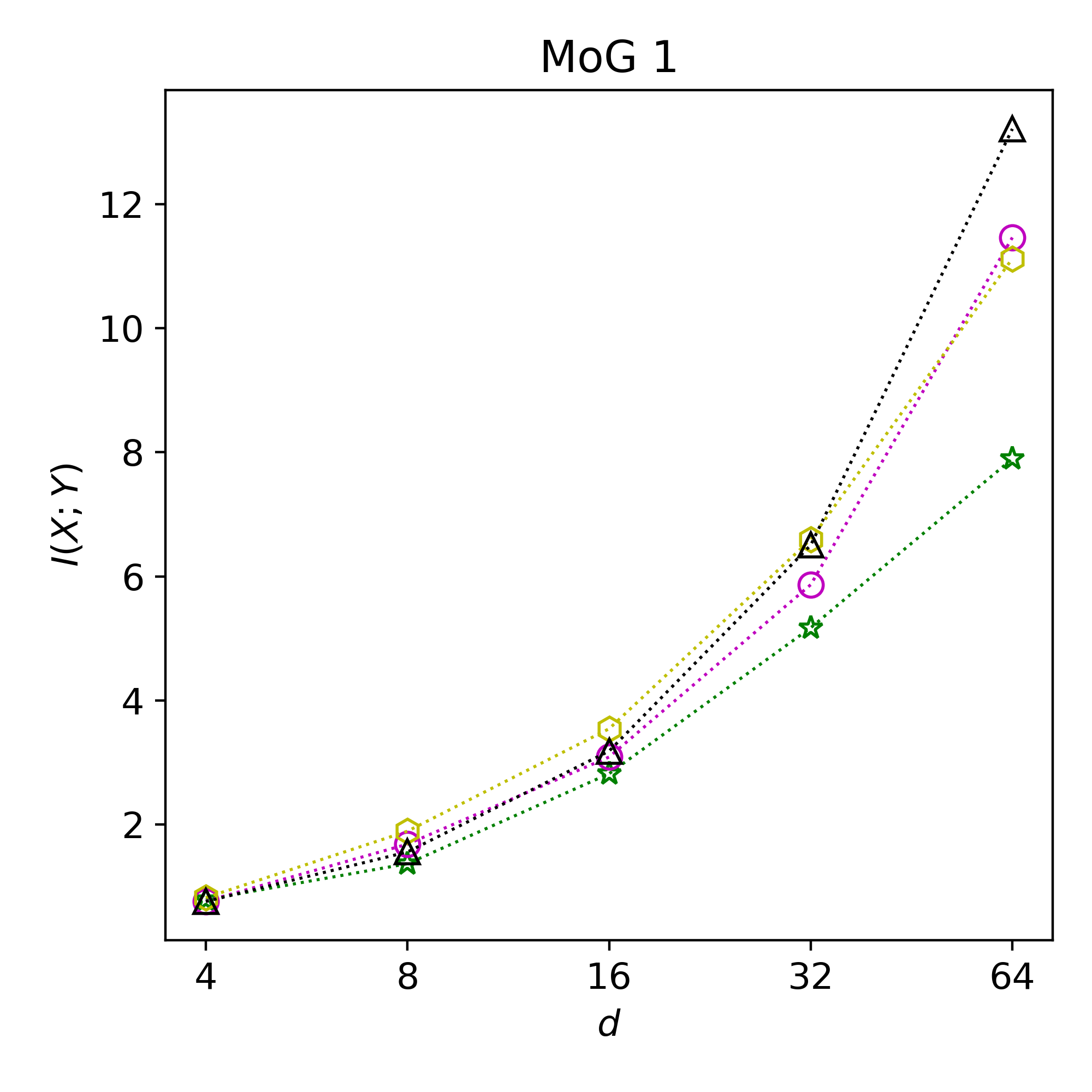}
\vspace{-8mm}
\caption{MoG 1}
\end{subfigure}
\hspace{-0.015\textwidth}
\begin{subfigure}{.255\textwidth}
\centering
\includegraphics[width=1.0\linewidth]{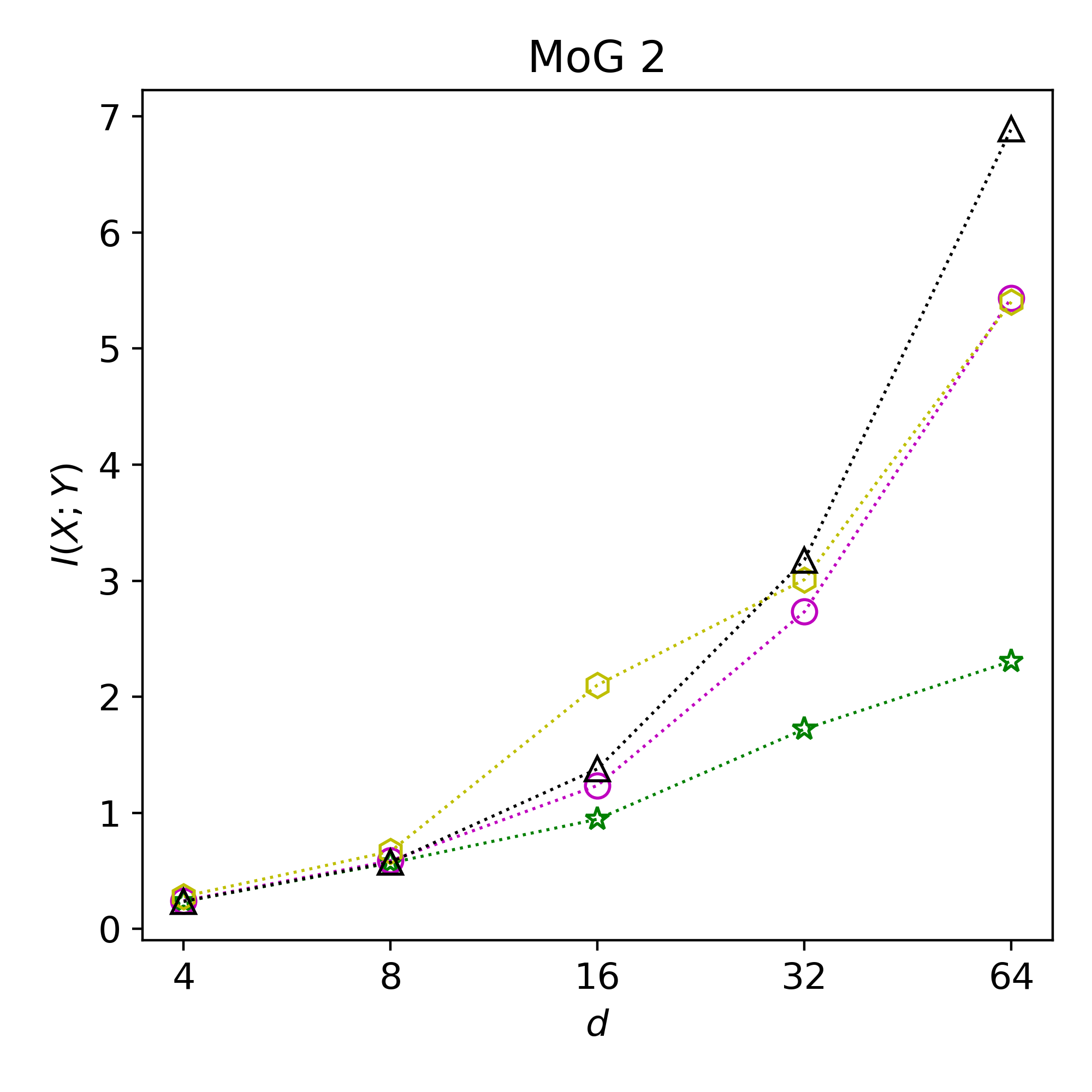}
\vspace{-8mm}
\caption{MoG 2}
\end{subfigure}
\caption{Comparison of different choices of reference distribution $q$. Here $N=10,000$.}
\label{fig:single_flow}
\end{figure}

\begin{figure}[!t]
\centering
\hspace{-0.03\textwidth}
\begin{subfigure}{.255\textwidth}
\centering
\includegraphics[width=1.0\linewidth]{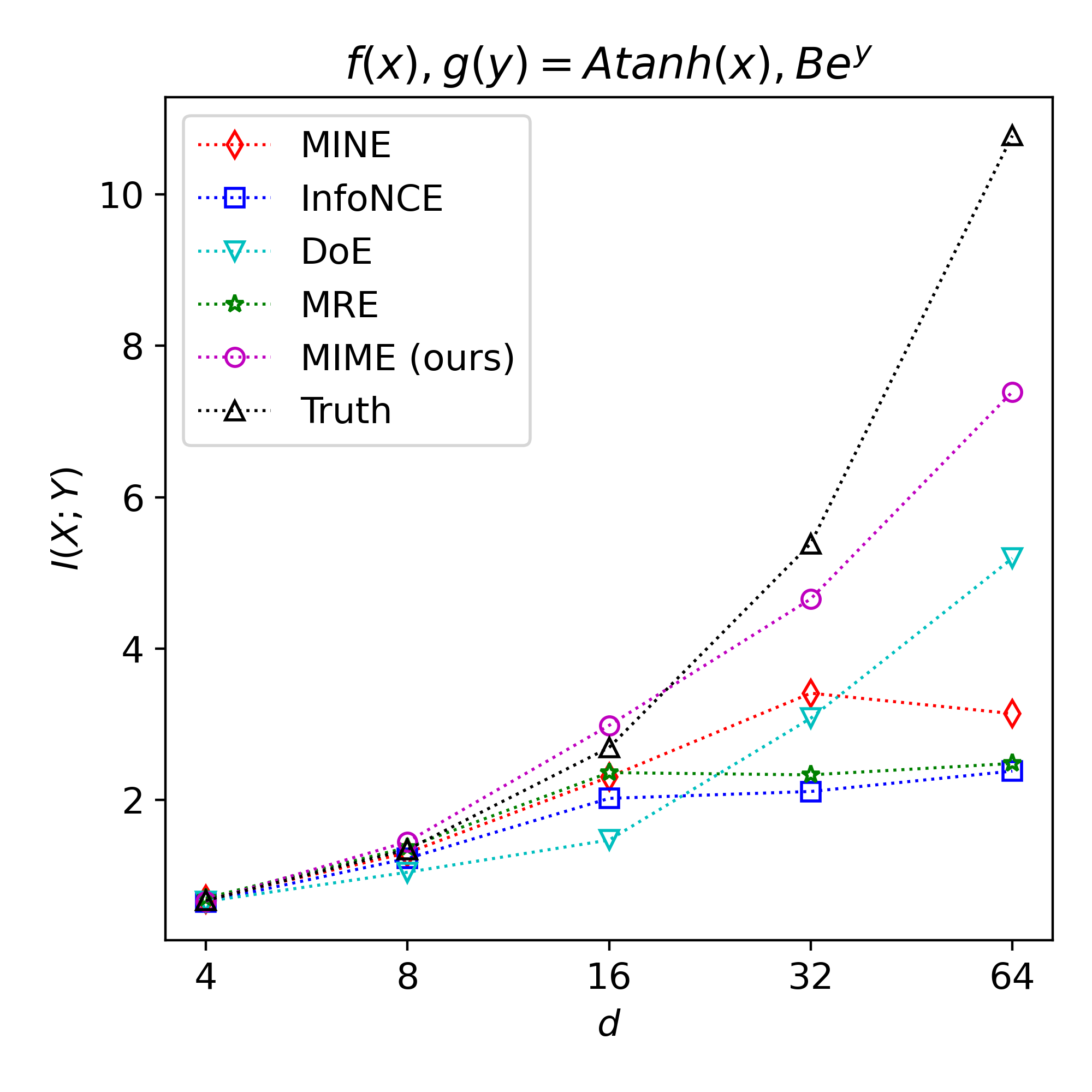}
\vspace{-8mm}
\caption{\centering $A\text{tanh}(x), Be^y$}
\end{subfigure}
\hspace{-0.015\textwidth}
\begin{subfigure}{.255\textwidth}
\centering
\includegraphics[width=1.0\linewidth]{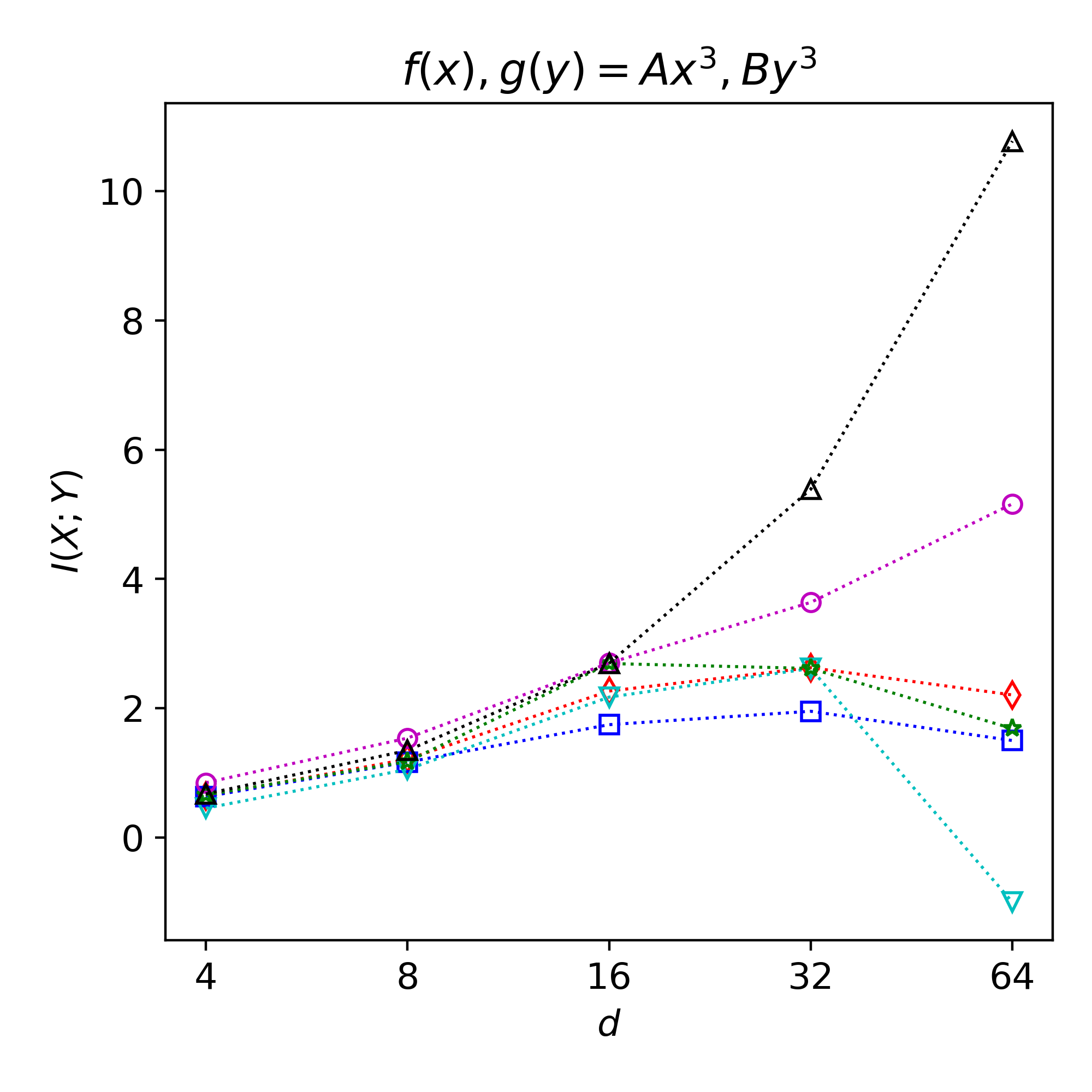}
\vspace{-8mm}
\caption{\centering $Ax^3, By^3$}
\end{subfigure}
\hspace{-0.015\textwidth}
\begin{subfigure}{.255\textwidth}
\centering
\includegraphics[width=1.0\linewidth]{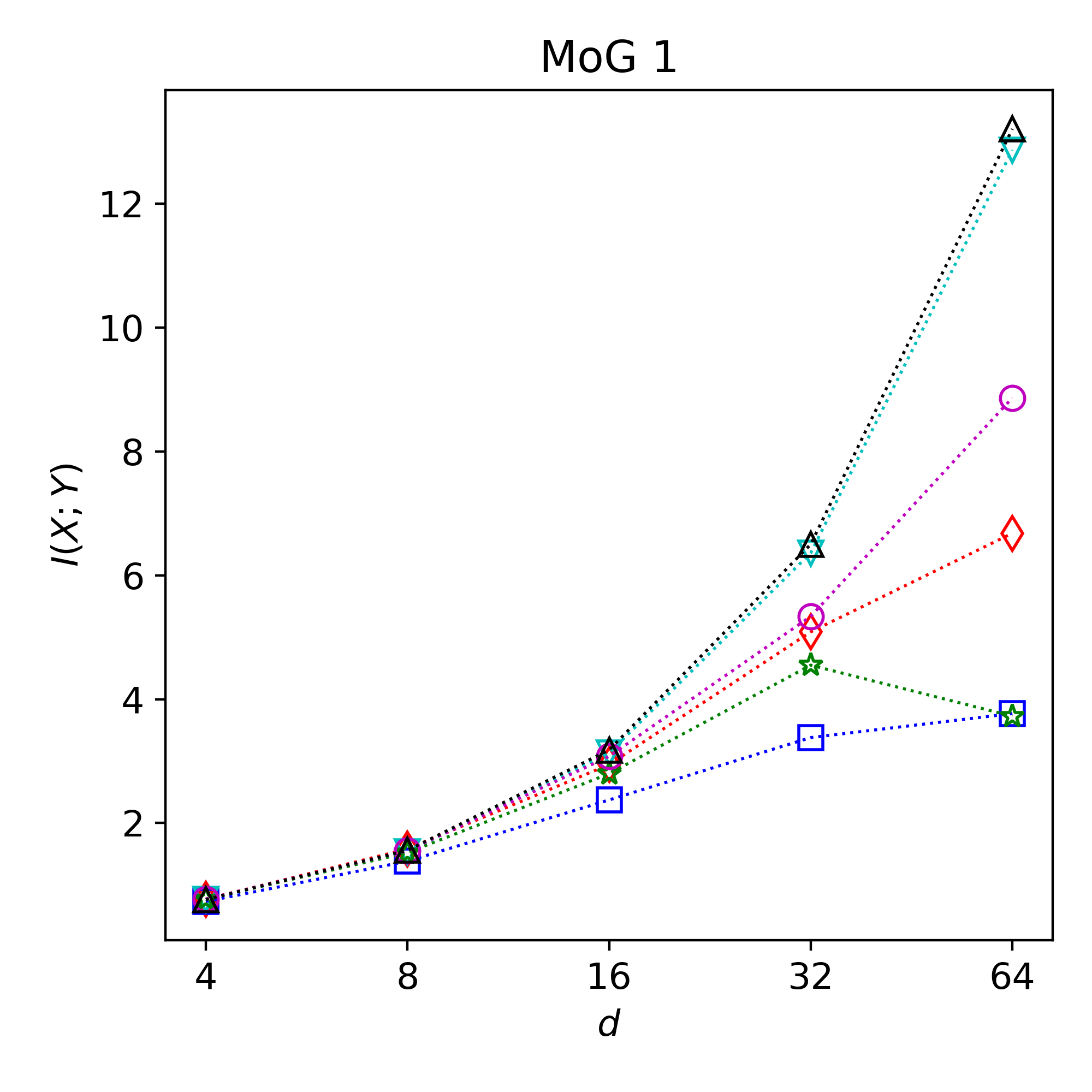}
\vspace{-8mm}
\caption{MoG 1}
\end{subfigure}
\hspace{-0.015\textwidth}
\begin{subfigure}{.255\textwidth}
\centering
\includegraphics[width=1.0\linewidth]{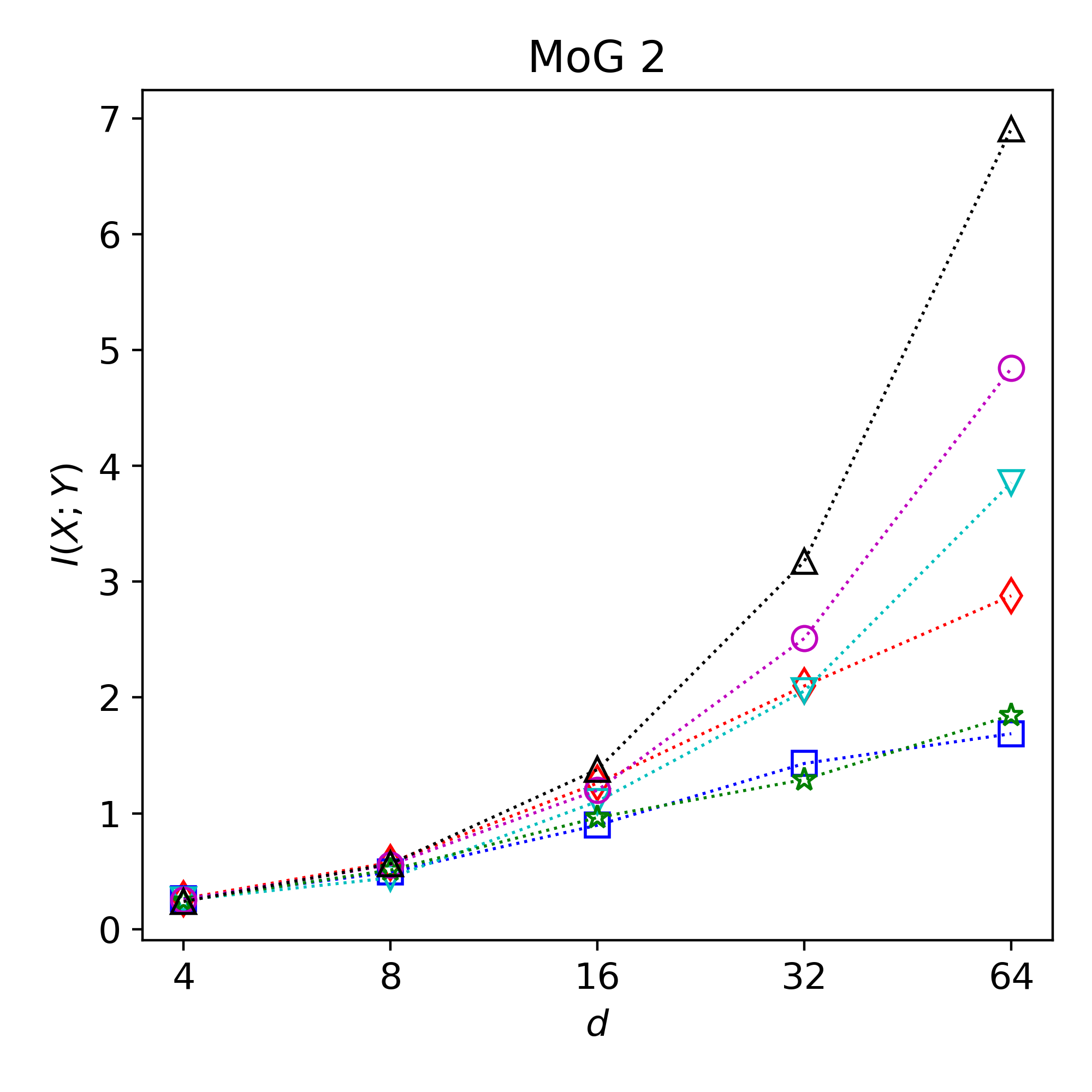}
\vspace{-8mm}
\caption{MoG 2}
\end{subfigure}
\caption{Comparison of different MI estimators when trained on $N =2,500$ data.}
\label{fig:small_N}
\end{figure}

\begin{figure}[!t]
\centering
\hspace{-0.03\textwidth}
\begin{subfigure}{.255\textwidth}
\centering
\includegraphics[width=1.0\linewidth]{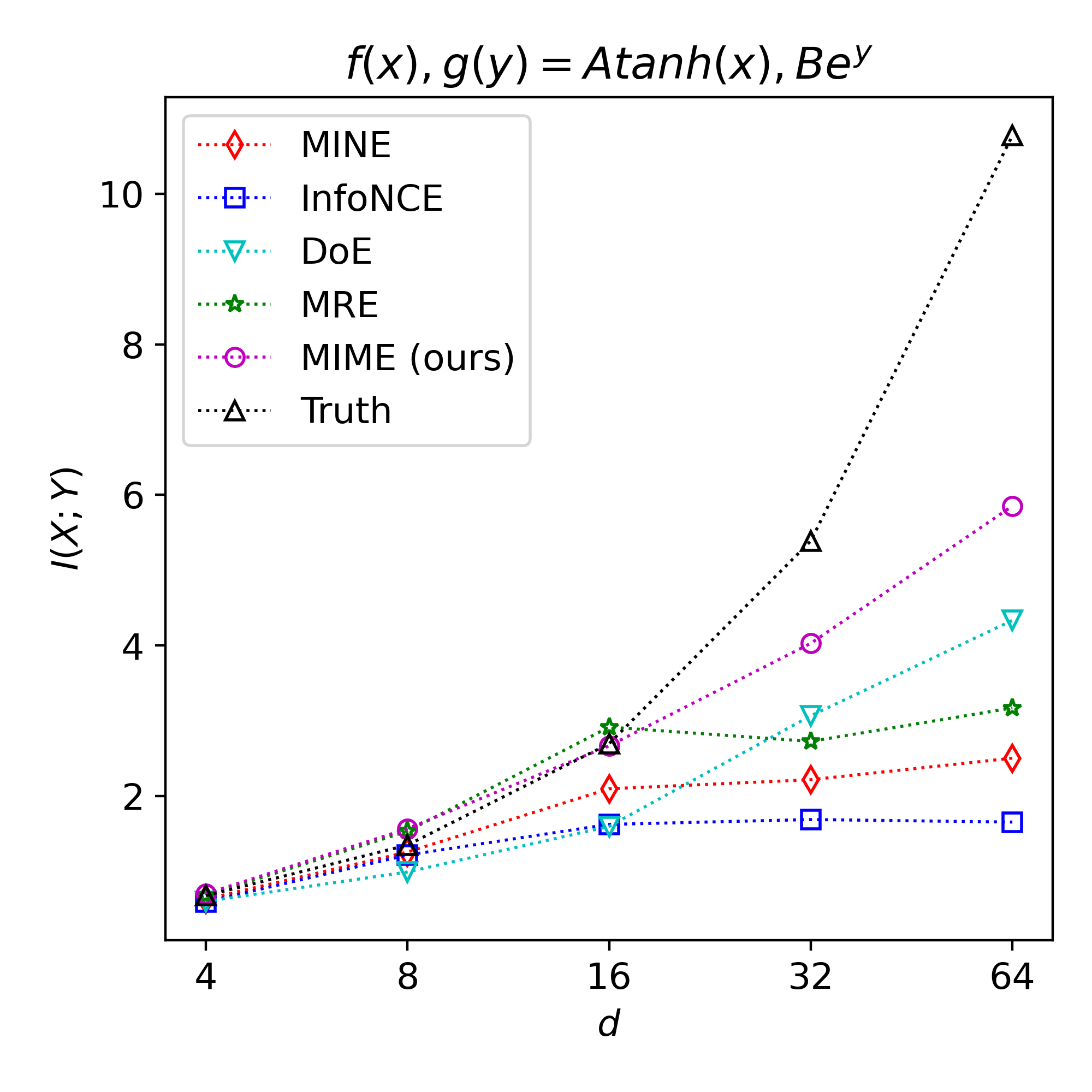}
\vspace{-8mm}
\caption{\centering $A\text{tanh}(x), Be^y$}
\end{subfigure}
\hspace{-0.015\textwidth}
\begin{subfigure}{.255\textwidth}
\centering
\includegraphics[width=1.0\linewidth]{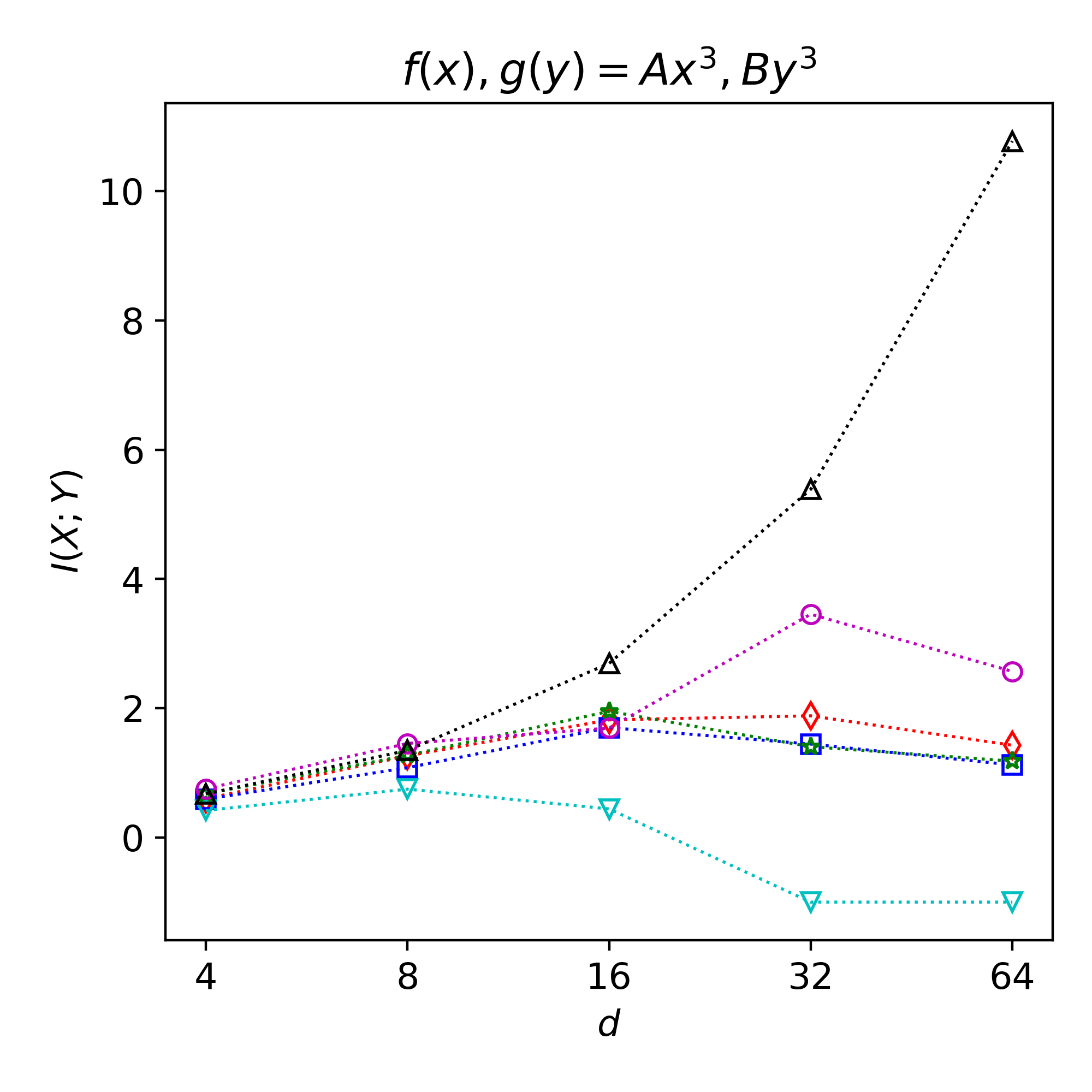}
\vspace{-8mm}
\caption{\centering $Ax^3, By^3$}
\end{subfigure}
\hspace{-0.015\textwidth}
\begin{subfigure}{.255\textwidth}
\centering
\includegraphics[width=1.0\linewidth]{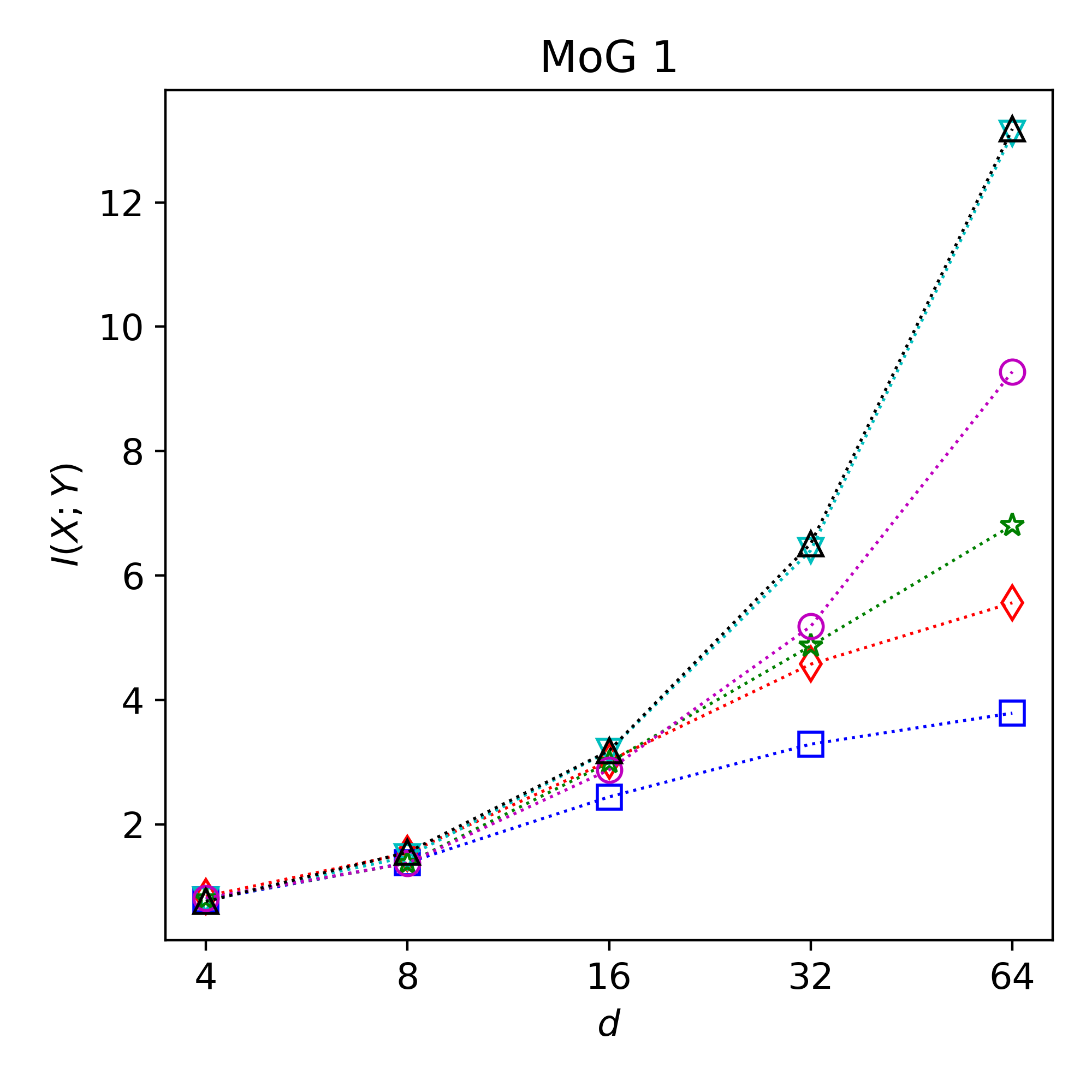}
\vspace{-8mm}
\caption{MoG 1}
\end{subfigure}
\hspace{-0.015\textwidth}
\begin{subfigure}{.255\textwidth}
\centering
\includegraphics[width=1.0\linewidth]{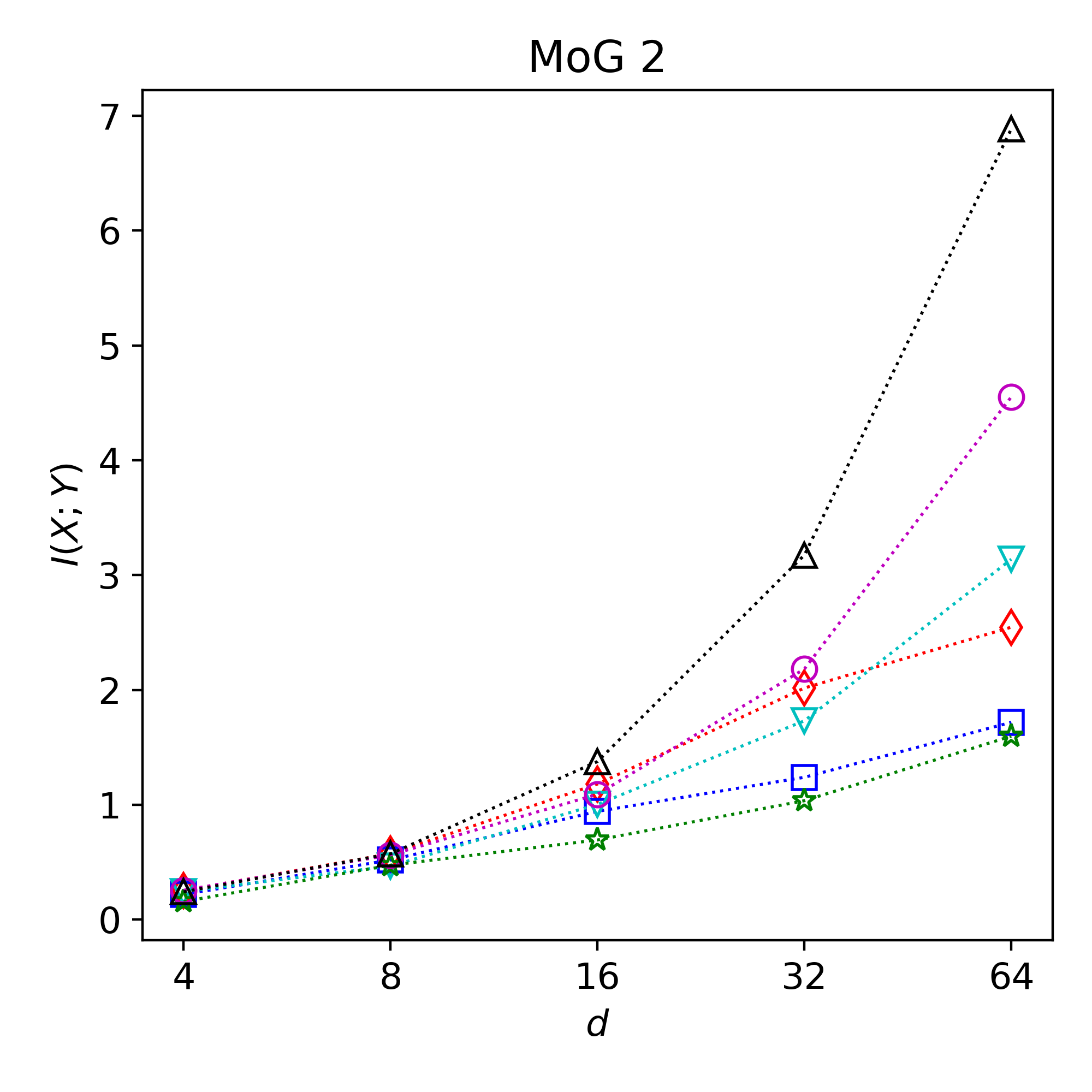}
\vspace{-8mm}
\caption{MoG 2}
\end{subfigure}
\caption{Comparison of different MI estimators when trained on $N =1,250$ data.}
\label{fig:tiny_N}
\end{figure}

\end{document}